\documentclass[preprint,12pt,3p]{elsarticle}

\usepackage{times}
\usepackage{epsfig}
\usepackage{graphicx}
\usepackage{amsmath}
\usepackage{amssymb}
\usepackage{amsthm}
\usepackage{dengxin}
\usepackage[table]{xcolor} 
\usepackage{xspace}
\usepackage[rightcaption]{sidecap}
\usepackage{enumerate}
\usepackage{multirow}
\usepackage[ruled, lined, linesnumbered] {algorithm2e}
\usepackage{array}
\usepackage{booktabs}

\def\dtrain{\mathcal{D}^\text{train}}

\journal{Pattern Recognition}

\begin{document}

\begin{frontmatter}

\title{Unsupervised High-level Feature Learning by Ensemble Projection for Semi-supervised Image Classification and Image Clustering
 \tnoteref{label0}}

\author{Dengxin Dai\corref{cor1}}
\ead{dai@vision.ee.ethz.ch}

\author{Luc Van Gool}
\ead{vangool@vision.ee.ethz.ch}

\address{Computer Vision Lab, ETH Z\"urich, CH-8092, Switzerland}
\cortext[cor1]{I am corresponding author}

\begin{abstract}
 This paper investigates the problem of image classification with limited or no annotations, but
 abundant unlabeled data. 
The setting exists in many tasks such as semi-supervised image classification, image clustering, and image retrieval.
Unlike previous methods, which develop or learn sophisticated regularizers for classifiers, our method learns a new
  image representation 
  by exploiting the distribution patterns of all available data. Particularly, a rich set of visual prototypes are sampled
  from all available data, and are taken as surrogate classes to train
  discriminative classifiers; images are projected  via
  the classifiers; the projected values, similarities to the
  prototypes, are stacked to build the new feature vector. The training
  set is noisy. Hence, in the spirit of ensemble learning we
  create a set of such training sets which are all diverse, leading to
 diverse classifiers. The method is dubbed Ensemble
  Projection (EP).  EP captures not only the characteristics of
  individual images, but also the relationships among images. It is
  conceptually simple and computationally efficient, yet effective and
  flexible.  Experiments on eight standard datasets show that: (1) EP
  outperforms previous methods for semi-supervised image
  classification; (2) EP produces promising results for self-taught
  image classification, where unlabeled samples are a random
  collection of images rather than being from the same distribution as
 the labeled ones; and (3) EP improves over the original features for
  image clustering. The code of the method is available at the project page.
\end{abstract}


\end{frontmatter}


\section{Introduction}
\label{sec:intro}
Providing efficient solutions to image classification has always been a
major focus in computer vision. Recent years have witnessed
considerable progress in image classification. However, most popular
systems \citep{lazebnik:cvpr06, Bosch:iccv07, Boiman_CVPR_2008, Indoor, siftllc:cvpr10, 
Sun_2010, Yang_2014_CVPR} heavily rely
on manually labeled training data, which is expensive and sometimes
impractical to acquire. Despite substantial efforts towards
efficient annotation by developing online games \citep{game:purpose}
or appealing software tools \citep{labelme}, collecting training data
for classification is still very time-consuming and tedious. The scarcity
of annotations, combined with the explosion of image data, starts
shifting focus towards learning with less supervision. As a result,
numerous techniques have been developed to learn classification models
with cheaper annotations. The most notable ones include unsupervised feature 
learning~\citep{stl-10, cnnfet14, feature:context, feature:LSTM},  
semi-supervised learning \citep{Fergus09, Guillaumin:cvpr:10, dai:iccv13b}, 
active learning \citep{JainK:cvpr09, cvpr09:multi:al},
transfer learning \citep{Transfer:CVPR:08, tl:survey},
weakly-supervised learning \citep{cvpr12:weak:video, metric:imitation}, self-taught learning
\citep{self-taught:icml07}, and image clustering
\citep{Sivic05b, dai}.

In this paper, we are interested in the problem of  image classification 
with limited or no annotation. Instead of regularizing the
classifiers like most of the previous methods \citep{SemiSVM, Joachims:1999,
  SemiBoost, SemiForest}, we learn a new feature representation 
using the all available data (labeled + unlabeled). 
Specifically, we aim to learn a new feature representation by exploiting 
the distribution patterns of the data to be handled. The setting assumes the availability of 
unlabeled data in the same or a similar distribution as the test data. 
This form of weak supervision is naturally available in applications such as
semi-supervised  image classification and image clustering, where data in the same or a similar 
distribution as the test data is available. The learned feature is specifically tuned for the data distribution of interest 
and performs better for the data than the standard features the method started with.  
The features to start with for our method can be hand-crafted features~\citep{gist, Ojala02,Bosch:iccv07,siftllc:cvpr10},  
learned features in a supervised manner~\cite{caffe14, rich:feature:cvpr14, deep:bmvc14} or 
learned features in an unsupervised way~\cite{stl-10, cnnfet14, feature:context, feature:LSTM, feature:video}.  

Learning with unlabeled data has been quite successful in many fields, for instance in semi-supervised learning 
(SSL)~\citep{Zhou:nips:04, deep:semi:embedding, Fergus09, SemiForest, Zhu:ISL:2009, 
nips14:ssl}, in image clustering~\cite{Grauman06, Frey_AffinityPropagation, dai}, and in 
unsupervised feature representation learning~\cite{stl-10, cnnfet14, feature:LSTM, feature:video}. 
Typically these methods build upon the
\emph{local-consistency} assumption that data samples with high
similarity should share the same label. This is also called
\emph{smoothness of manifold}, and it is often used to regularize the training process for the classifiers or feature
 representations.  
In this paper, we propose another way to exploit the \emph{local-consistency} assumption to 
learn a new feature representation. 
The new feature
representation is learned in a discriminative way to capture not only
the information of individual images, but also the relationships among
images. The learning is conceptually straightforward and
computationally simple. The learned features can be fed into any
classifiers for the final classification of the unlabeled samples.
Thus, the method is agnostic to the classifier choice. This facilitates the
deployment of SSL methods, as users often have their favorite
classifiers and are reluctant to drop them. For image clustering, we
apply the same feature learning methods to all provided images, and
then feed the learned features to standard clustering methods such as
$k$-means and Spectral Clustering.  Below, we present our motivations
and outline the method.

\begin{figure*}[t]
  \centering
  \includegraphics[width=0.98\textwidth]{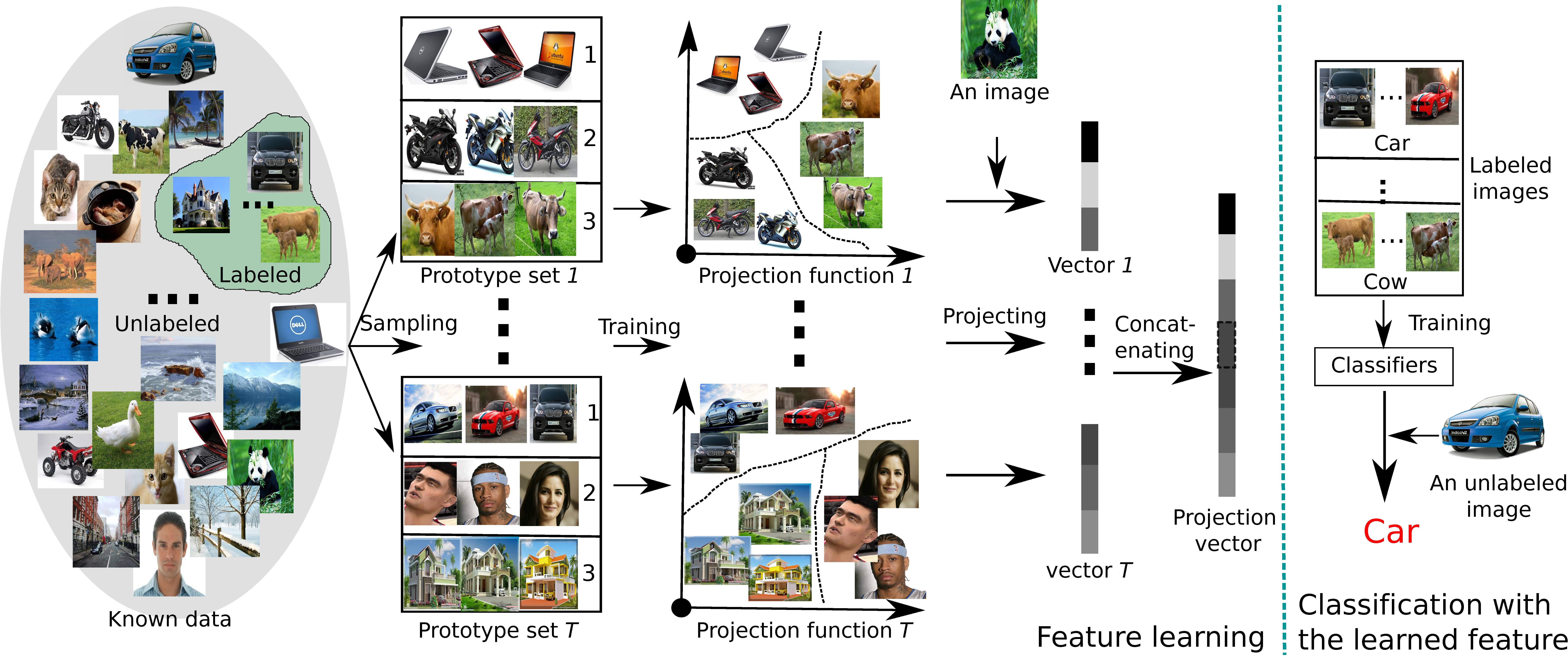}
  \caption{The pipeline of Ensemble Projection (EP). EP consists of
    unsupervised feature learning (left panel) and plain
    classification or clustering (right panel). For feature learning,
    EP samples an ensemble of $T$ diverse prototype sets from all
    known images and learns discriminative classifiers on them for the
    projection functions. Images are then projected using these
    functions to obtain their new representation. These features are
    fed into standard classifiers and clustering methods for image
    classification and clustering respectively.}
\label{fig:pipeline}
\end{figure*}

\subsection{Motivations}
People learn and abstract the concepts of object classes well from
their intrinsic characteristics, such as colors, textures, and
shapes. For instance, \emph{sky} is blue, and a \emph{football} is
spherical. We also do so by comparing new object classes to those classes that
have already been learned. For example, a \emph{leopard} is similar in
appearance to a \emph{jaguar}, but is smaller. This paradigm of
learning-by-comparison or characterization-by-comparison is part of
Eleanor Rosch's prototype theory \citep{Rosch:1978}, that states that
an object's class is determined by its \emph{similarity} to prototypes
which represent object classes. The theory has been used successfully
in transfer learning \citep{Transfer:CVPR:08}, when labeled data of
different classes are available. An important question is whether the
theory can also be used for feature representation learning when a large
amount of unlabeled data is available. This paper investigates this
problem.

To use this paradigm, we first need to create the prototypes
automatically from the available data. In
keeping with Eleanor Rosch's prototype theory \citep{Rosch:1978}, ideal
prototypes should have two properties: 1) images in the same prototype
are to be from the same class; and 2) images of different
prototypes are to be from different classes. They 
guarantee meaningful comparisons and reduce ambiguity.  Without
access to labels of data samples, the prototypes have to be created in
an unsupervised way, based on some assumptions. In addition to the
widely-used \emph{local-consistency}, we propose another one called
\emph{exotic-consistency}, which states that samples that are far
apart in the feature space are very likely to come from
different classes. The assumptions have been verified
experimentally, and will be presented in Section~\ref{sec:mov1}.  Based
on these two assumptions, it stands to reason that samples along with
their closest neighbors can be ``good'' prototypes, and  such
prototypes that are far apart can play the role of different classes.  According to
this observation, we design a method to sample the prototype set from
all available images by encoding them on a graph with links
reflecting their affinity.

The sampled prototypes are taken as surrogate classes and
discriminative learning is yields projection functions tuned to the
classes. Images are then linked to the prototypes via their projection
values (classification scores) by the functions. Since information
carried by a single prototype set is limited and can be noisy, we
borrow ideas from ensemble learning \citep{EnClasReview} to create an
ensemble of diverse prototype sets, which in turn leads to an ensemble
of projection functions, to mitigate the influence of the deficiencies
of each training set. The idea is that if the deficiency modes of the
individual training sets are different or `orthogonal', ensemble
learning is able to cancel out or at least mitigate their effect. This
conjecture is verified with a simulated experiment in
Section~\ref{sec:mov2}, and is also supported by the superior performance
of our method in real applications. With the ensemble of classifiers,
images are then represented by the concatenation of their
classification scores -- similarities to all the sampled image
prototypes -- for the final classification, which is in keeping with
prototype theory \citep{Rosch:1978}. We call the method Ensemble
Projection (EP). Its schematic diagram is sketched in
Figure \ref{fig:pipeline}.

\subsection{Contributions}
EP was evaluated on eight image classification datasets, ranging from
texture classification, over object classification and scene
classification, to style classification. For SSL, EP is compared
to three baselines and three other methods. For image clustering, EP is
compared to the original features it started with. Two standard clustering methods are used: $k$-means and
Spectral Clustering.  The experiments show that: (1) EP improves over
the original features by exploiting the data distribution of interest, 
and outperforms competing SSL methods; (2) EP
produces promising results for self-taught image classification where
the unlabeled data does not follow the same distribution as the
labeled ones; (3) EP improves over the original features for image clustering.

This paper is an extension of our conference papers~\citep{dai:eccv12b,dai:iccv13b}. 
In addition to putting the two tasks, image clustering and semi-supervised image classification, 
into the same framework, this paper brings several new
contributions. First, in the conference papers, EP
was validated only with hand-crafted features, such as LBP
\citep{Ojala02}, GIST \citep{gist}, and PHOG
\citep{Bosch:iccv07}. These features, however, are obsolete for image classification. 
Recently, features learned by CNN has resulted in
state-of-the-art performance in various classification tasks
\citep{nips12:cnn, caffe14, rich:feature:cvpr14, deep:bmvc14}. In this
paper, we validate the efficacy of EP also with CNN features. Second, experiments are 
conducted on eight standard classification
datasets instead of only four in \citep{dai:iccv13b}. Third, 
more analyses and insights are given. 
Our feature learning method can be used for other tasks as well. For instance, 
\citep{random:hashing} extended the idea to generate hashing functions
for efficient image retrieval.

The rest of this paper is organized as follows. Section~\ref{sec:related}
reports on related work. Section~\ref{sec:observations} describes the
observations that motivate the method. Section~\ref{sec:approach} is devoted to the approach, followed by experiments in
Section~\ref{sec:experiments}. Section~\ref{sec:conclusion} concludes the
paper.

\section{Related Work}
\label{sec:related}
Our method is generally relevant to image feature learning, semi-supervised
learning, ensemble learning, and image clustering.

\textbf{Supervised Feature Learning}: Over the past years, a wide spectrum of
features, from pixel-level to semantic-level, have been designed and
used for different vision tasks. Due to the semantic gap, recent work
extract high-level features, which go beyond single images and are
probably impregnated with semantic information. Notable examples are
Image Attributes~\citep{ObjectAttribute:cvpr09},
Classemes~\citep{eccv10:classemes}, and Object
Bank~\citep{li:objectbank}. While getting pleasing results, these
methods all require additional labeled training data, which is exactly
what we want to avoid.  There have been
attempts, \eg~\citep{augmented_attribute:eccv12, design_attribute:cvpr13}, to
avoid the extra attribute-level supervision, but they still require
canonical class-level supervision. Our representation learning
however, is fully unsupervised.  
The pre-trained CNN features~\citep{nips12:cnn, caffe14, rich:feature:cvpr14, deep:bmvc14}
have shown state-of-the-art performance on various classification tasks. Our feature 
learning is complementary to their methods. As shown in the experiment, our method can improve on top of 
the CNN features by exploiting the distribution patterns of the data to be classified.  
Although the technique of fine-tuning can 
boost the performance of CNN features for 
the specific tasks at hand~\citep{midlevel:transfer, cnn:transferable}, it needs labeled data of a moderate size, 
which is not always available in our setting.
Our method can be understood as unsupervised feature enhancing or fine-tuning.

\textbf{Unsupervised Feature Learning}: Our method is akin to methods which learn middle- or high-level image
representation in an unsupervised manner. \citep{stl-10} employs
$k$-means mining filters of image patches and then applys the filters
for feature computation. \citep{cnnfet14} generates surrogate classes
by augmenting each patch with its transformed versions under a set of
transformations such as translation, scaling, and rotation, and trains
a CNN on top of these surrogate classes to generate features. The idea
is very similar to ours, but our surrogate classes are generated by
augmenting seed images with their close neighbors. The learning
methods are also different. \citep{mid-level:patches} discovers a set
of representative patches by training discriminative classifiers with
small, compact patch clusters from one dataset, and testing them on
another dataset to find similar patches. The found patches are then
used to train new classifiers, which are applied back to the first
dataset. The process iterates and terminates after rounds, resulting
in a set of representative patches and their corresponding
`filters'. The idea of learning `filters' from compact clusters shares
similarities with what we do, but our clusters are images rather than
patches. Other forms of weak supervision have also been exploited to learn good 
feature representation without human labeled data, and they all obtain very promising results. 
For instance, \citep{feature:context} uses the spatial relationships of 
image windows in an image as the supervision to train a neural network; \citep{feature:video} exploits 
the tracking results of objects in videos to guide the training of a neural network to learn 
feature representations; and \citep{learning:by:moving} exploits the ego-motion of cameras for the training.
These methods aim for general feature representation. Our method, however, is designed to 
`tune' or enhance vision features specifically for the datasets on which the vision tasks are performed.  

\textbf{Semi-supervised Learning}: SSL aims at enhancing the
performance of classification systems by exploiting an additional set of
unlabeled data. Due to its great practical value, SSL has a rich
literature~\citep{book06:ssl, Zhu:ISL:2009}. Amongst existing methods,
the simplest methodology for SSL is based on the self-training scheme
\citep{co-training:98} where the system iterates between training
classification models with current `labeled' training data and augmenting
the training set by adding its highly confident predictions in the set
of unlabeled data; the process starts from human labeled data and
stops until some termination condition is reached, \eg the maximum
number of iterations.  \citep{Guillaumin:cvpr:10} and \citep{Semi:eccv12} presented two methods in this stream for image classification. While
obtaining promising results, they both require additional supervision:
\citep{Guillaumin:cvpr:10} need image tags and \citep{Semi:eccv12}
image attributes.

The second group of SSL methods is based on label propagation over a
graph, where nodes represent data examples and edges reflect their
similarities. The optimal labels are those that are maximally
consistent with the supervised class labels and the graph
structure. Well known examples include Harmonic-Function
\citep{Zhu:Harmonic:03}, Local-Global Consistency \citep{Zhou:nips:04},
Manifold Regularization \citep{Belkin:semiframe:2006}, and
Eigenfunctions \citep{Fergus09}. While having strong theoretical
support, these methods are unable to exploit the power of
discriminative learning for image classification.

Another group of methods utilize the unlabeled data to regularize the
classifying functions -- enforcing the boundaries to pass through
regions with a low density of data samples. The most notable methods
are transductive SVMs \citep{Joachims:1999}, Semi-supervised
SVMs \citep{SemiSVM}, and semi-supervised random
forests \citep{SemiForest}. These methods have difficulties to extend to
large-scale applications, and developing an efficient optimization for
them is still an open question.  Readers are referred
to \citep{Zhu:ISL:2009} for a thorough overview of SSL.


\textbf{Ensemble Learning}: Our method learns the representation from
an ensemble of prototype sets, thus sharing ideas with ensemble
learning (EL). EL builds a committee of base learners, and finds
solutions by maximizing the agreement. Popular ensemble methods that
have been extended to semi-supervised scenarios are
Boosting~\citep{SemiBoost} and Random
Forests~\citep{SemiForest}. However, these methods still differ
significantly from ours. They focus on the problem of improving
classifiers by using unlabeled data. Our method learns new
representations for images using all data available. Thus, it is
independent of the classification method. The reason we use EL is to
capture rich visual attributes from a series of prototype sets, and to
mitigate the deficiency of the sampled prototype sets. Other work
close to ours is 
Random Ensemble Metrics~\citep{ensemble:iccv11}, where images are
projected to randomly subsampled training classes for supervised
distance learning.

\textbf{Image Clustering}: A plethora of methods have been developed for image clustering.  
\citep{Fergus03} modeled
objects as constellations of visual parts and estimated parameters
using the expectation-maximization algorithm for unsupervised
classification. \citep{Sivic05b} proposed using aspect models
to discover object classes from an unordered image
collection. Later on, \citep{Sivic08} used Hierarchical Latent
Dirichlet Allocation to automatically discover object class
hierarchies. For scene class discovery, \citep{dai}
proposed to combine information projection and clustering
sampling. These methods assume explicit distributions for the samples.
Image classes, nevertheless, are arranged in complex
and widely diverging shapes, making the design of explicit models
difficult. 
An alternative strand, which is more versatile in handling structured
data, builds on similarity-based methods. 
\citep{Frey_AffinityPropagation} applied the affinity propagation
algorithm of \citep{Frey_Dueck_2007} for unsupervised image
categorization.  \citep{Grauman06} developed
partially matching image features to compute image similarity and used
spectral methods for image clustering.
The main difficulty of this strand is how to measure image similarity
as the semantic level goes up.  
Readers are referred to \citep{Tuytelaars_UnsupervisedSurvey} for a survey.



\section{Observations}
\label{sec:observations}

\label{sec:observation} In this section, we motivate our approach and explain why it is
working. We experimentally verify our assumptions: First, given a
standard distance metric over images, do the assumptions
\emph{local-con\-sist\-ency} and \emph{exotic-con\-sistency}
hold, and to what extent? Second, is ensemble learning able to cancel
out the deficiency of the individual training sets, given that the
number of such training sets are sufficiently large and the deficiency
modes of them are different or `orthogonal'?

\subsection{Observation 1}
\label{sec:mov1}
The assumptions of \emph{local-consistency} and
\emph{exotic-consist\-ency} do hold for real image datasets.  An
ideal image representation along with a distance metric should ensure
that all images of the same class are more similar to each other
than to those of other classes. However, this does not strictly hold for
most of vision systems in reality. In this section, we want to verify
whether the relaxed assumptions 
\emph{local-consistency} and the \emph{exotic-consistency} hold. 
These state images are very likely from the same class as their close
neighbors, and very likely from different classes than those far from them. 
In order to examine the assumptions, we tabulate how often an
image is from the same class as its $k^{th}$-nearest neighbor.  We
refer to the frequency as label co-occurrence probability $p(k)$.
$p(k)$ is averaged across images and class labels in the dataset. Four
features were tested: GIST \citep{gist}, PHOG~\citep{Bosch:iccv07},
LBP \citep{Ojala02}, and the CNN feature~\citep{deep:bmvc14}. 
The Euclidean distance is used here.

\begin{figure*}[ht]
$ \begin{array}{cccc} \hspace{-2mm}
\includegraphics[width=0.33\textwidth]{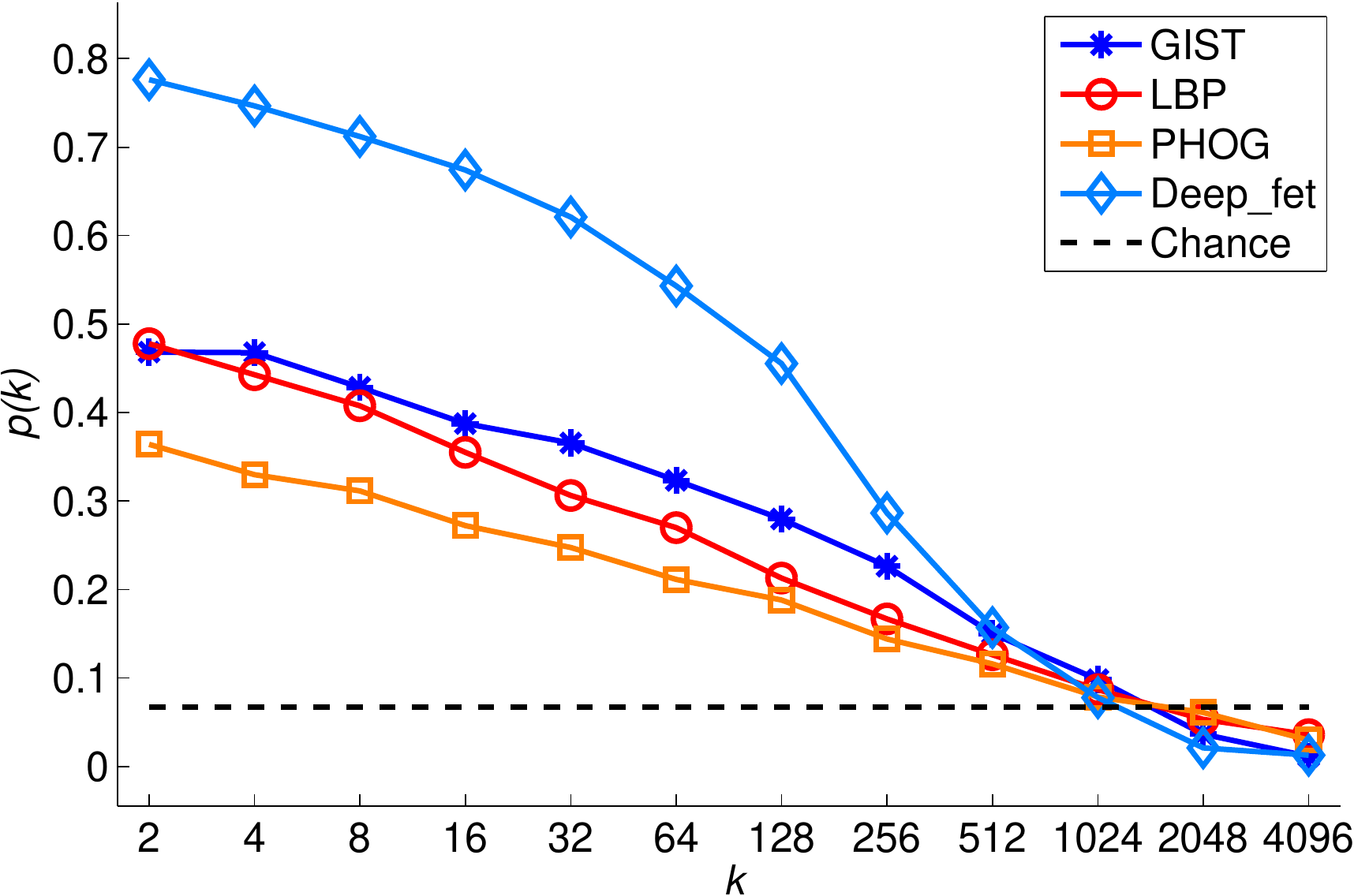} & \hspace{-3mm}
\includegraphics[width=0.33\textwidth]{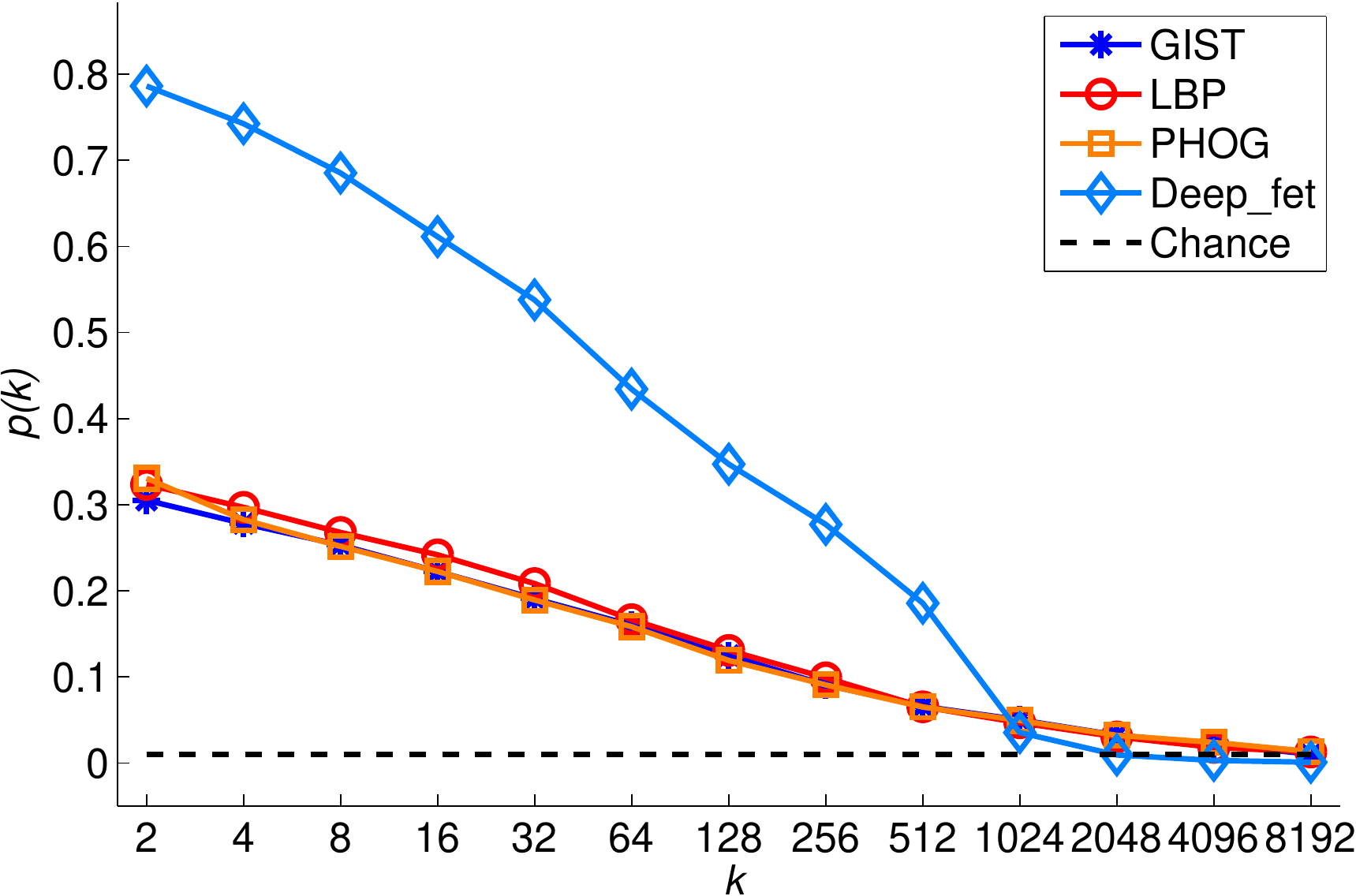} & \hspace{-3mm} 
\includegraphics[width=0.33\textwidth]{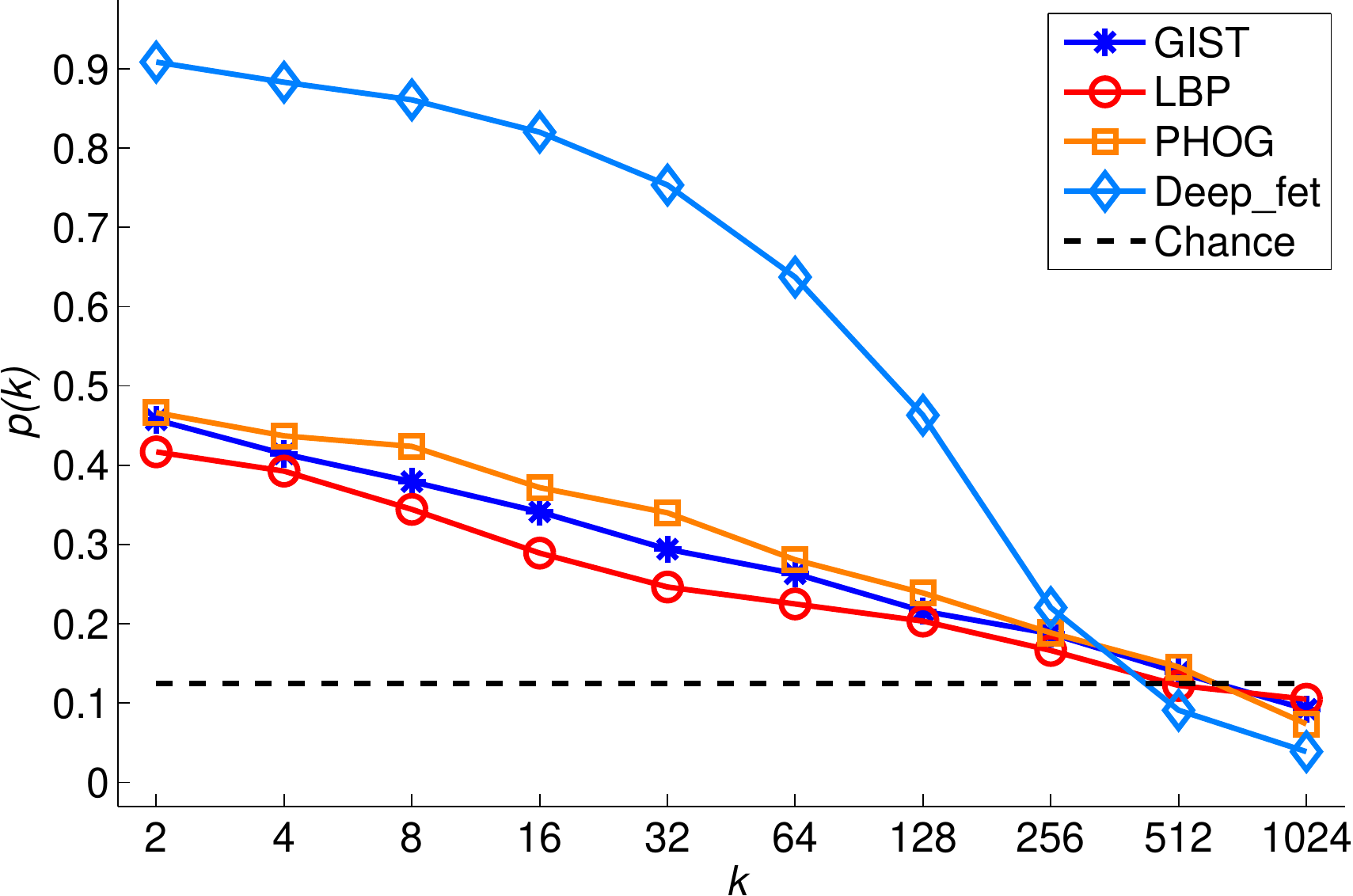} \\ 
\scriptsize{\text{(a) Scene-15 (15,4485) }}  & \scriptsize{\text{(b) Caltech-101 (101, 8677)} }  & \scriptsize{\text{(c) Event-8 (8, 1574)} }  \\

\includegraphics[width=0.33\textwidth]{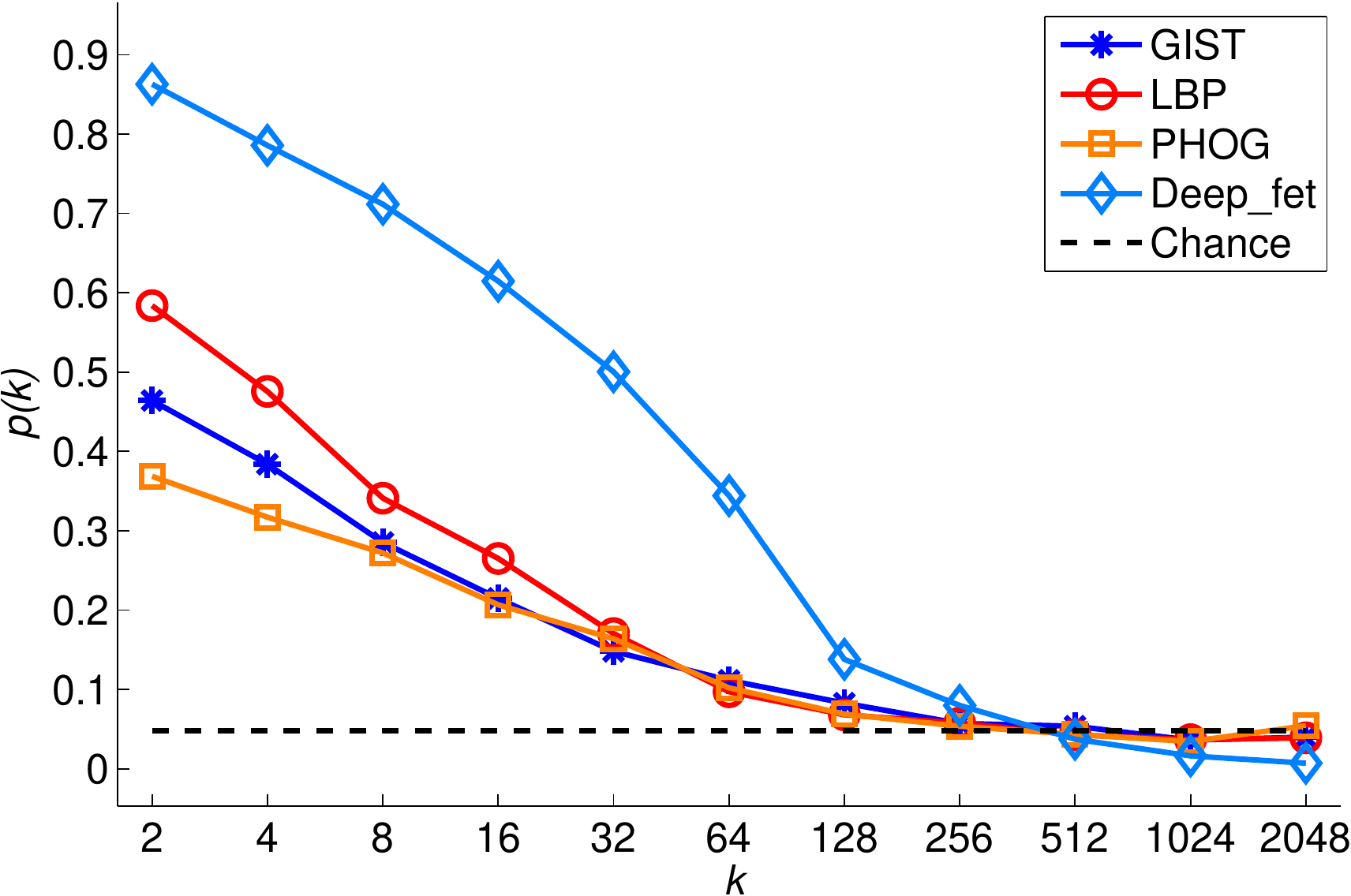} & \hspace{-3mm}
\includegraphics[width=0.33\textwidth]{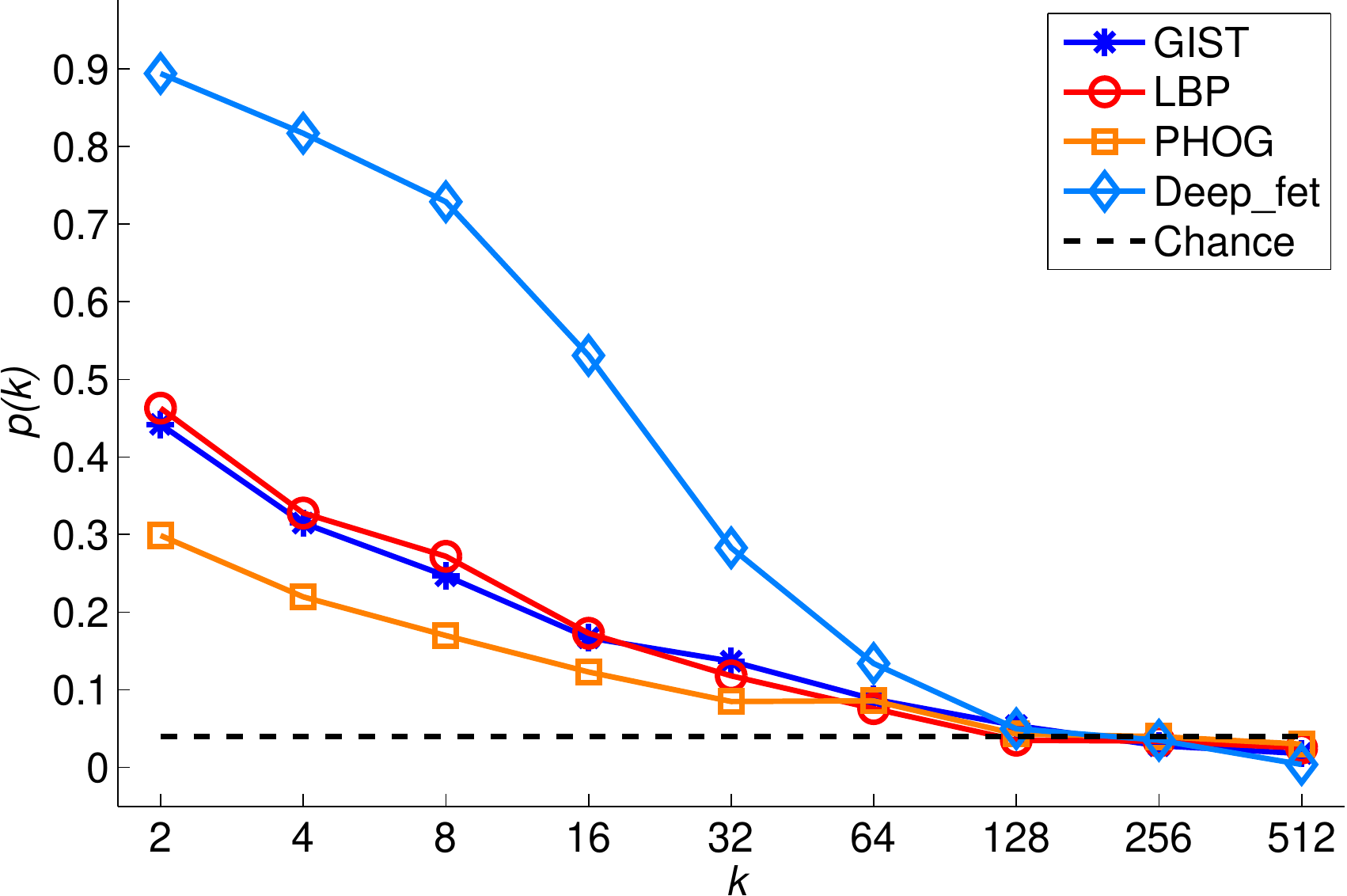} & \hspace{-3mm} 
\includegraphics[width=0.33\textwidth]{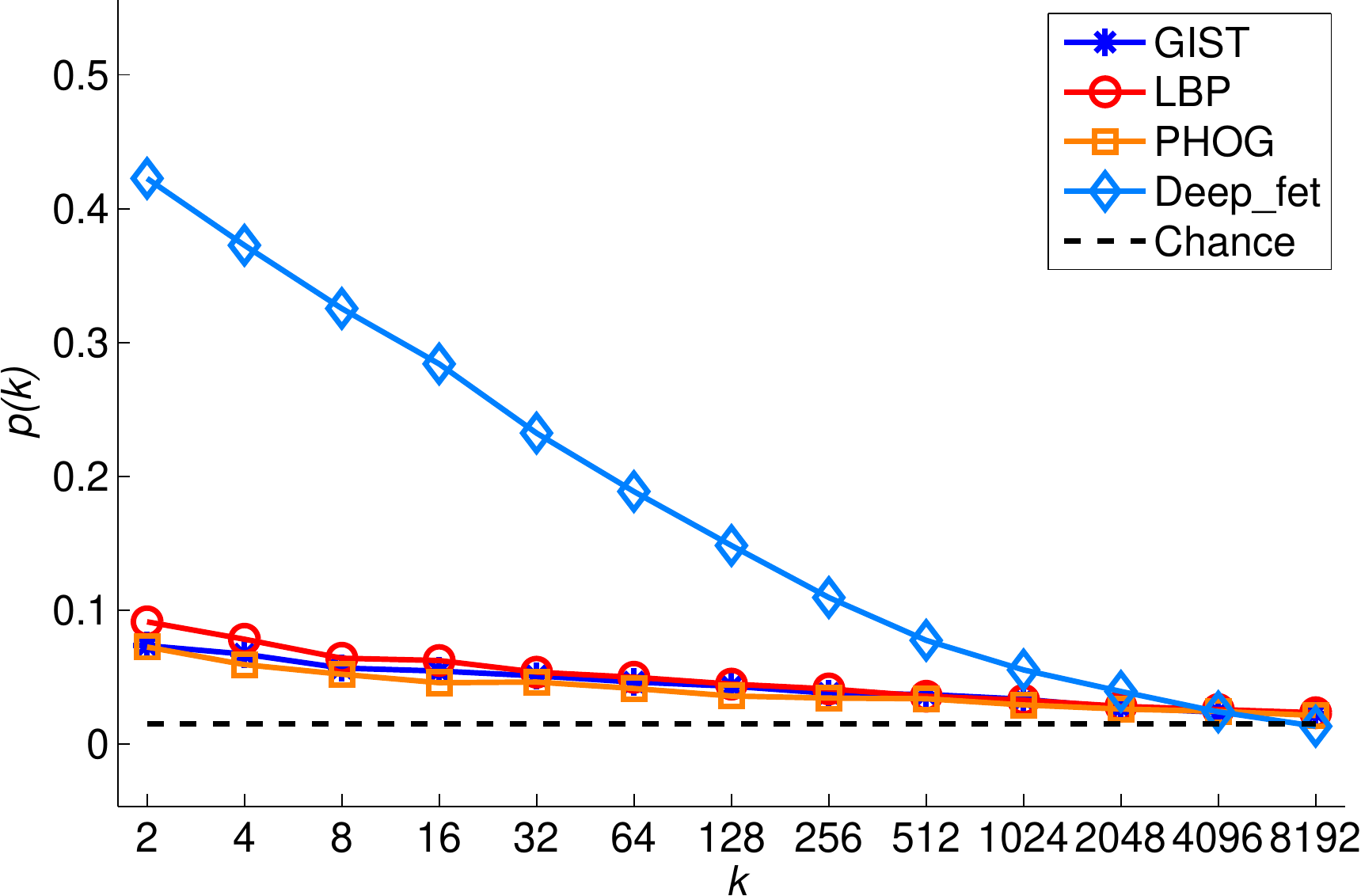} \\ 
\scriptsize{\text{(d) LandUse-21 (21, 2100)} }  & \scriptsize{\text{(e) Texture-25 (25, 1000)} }  & \scriptsize{\text{(f) Indoor-67 (67, 15613)} }  \\
\end{array} $ 
\caption{The label co-occurrence probability $p(k)$: frequency of
  images having the same label with their $k_{th}$ neighbors.  Results
  on six datasets are shown. The number of classes and the number
  of images of the datasets are shown as well. }
\label{fig:mov1}
\end{figure*}

Figure~\ref{fig:mov1} shows the results on six datasets (Datasets and
features will be introduced in Section~\ref{sec:experiments}). The results
reveal that using the distance metric in conventional ways (\eg
clustering by $k$-means and spectral methods) will result in very noisy
training sets, because the label co-occurrence probability $p(k)$
drops very quickly with $k$.  Sampling in the very close
neighborhood of a given image is likely to generate more instances of
the same class, whereas sampling far-away tends to gather
samples of different classes.  This suggests that samples along with a
few very close neighbors, namely ``compact'' image clusters, can form
a training set for a single class, and a set of such image clusters
far away from each other in feature space can serve as 
good prototype sets for different classes. Furthermore, sampling in
this way provides the chance of creating a large number of diverse
prototype sets, due to the small size of each sampled prototype set.
Also, from this figure, it is evident that the CNN feature performs
significantly better than the rest, which suggests
that using the CNN feature in our system is recommendable.

\begin{figure*}[ht]
  \centering
$  \begin{array}{cccc}
 \includegraphics[width=0.47\textwidth, height=55mm]{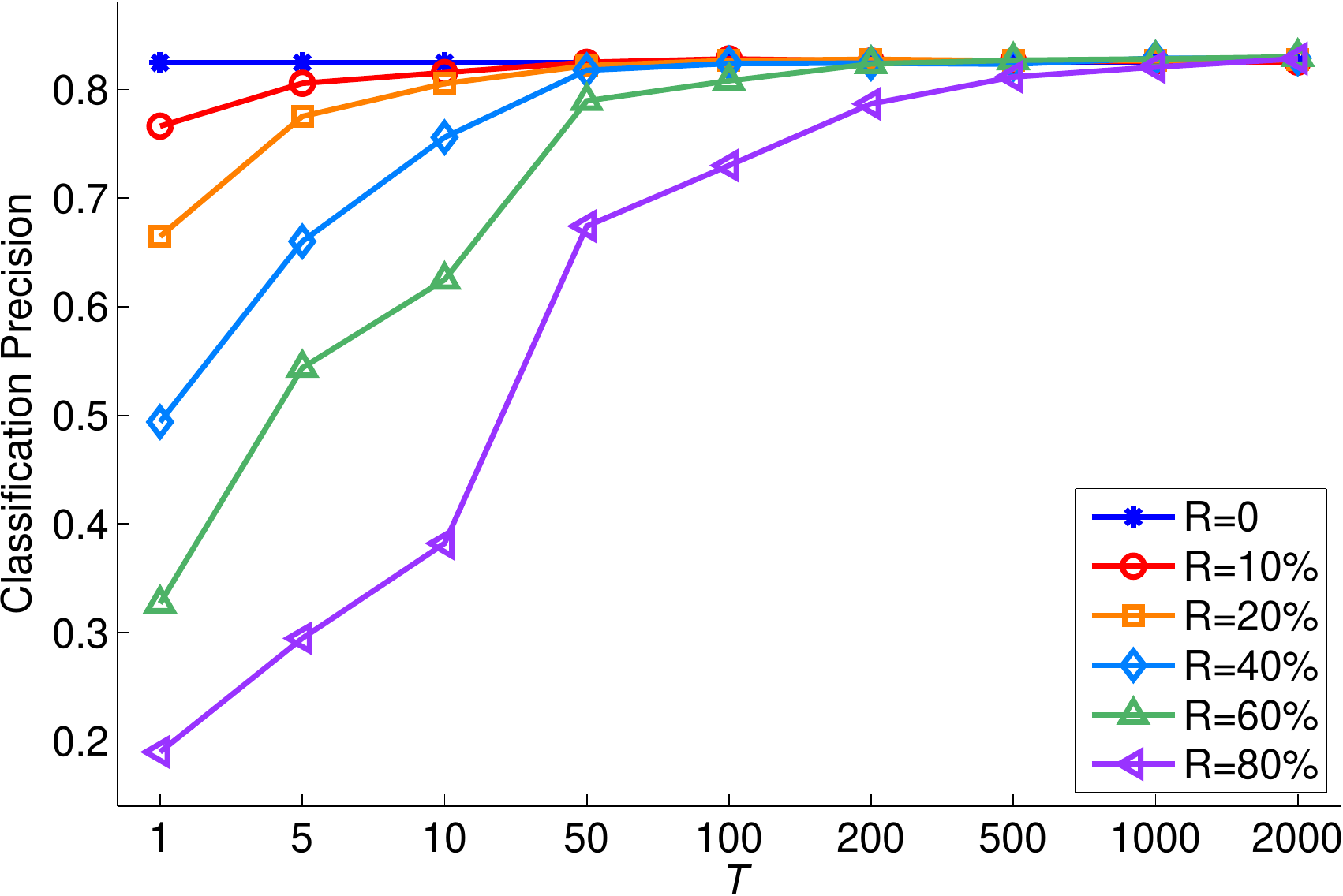} & 
  \includegraphics[width=0.47\textwidth, height=55mm]{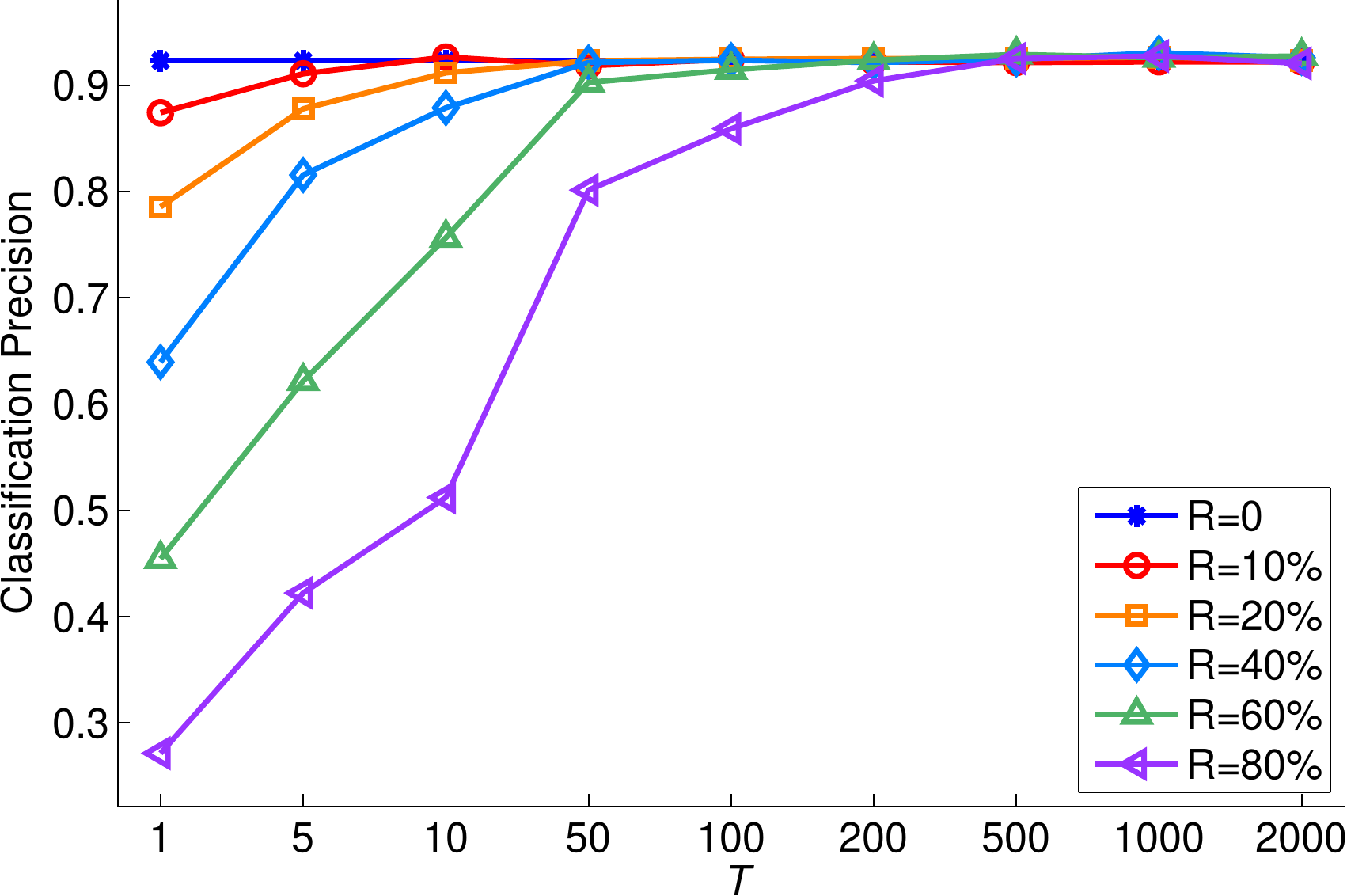}  \\
\scriptsize{\text{(a) Scene-15}}  & \scriptsize{\text{(b) Event-8}}
  \end{array}  $
  \caption{Classification accuracy of ensemble learning on the the
    Scene-15 dataset~\citep{lazebnik:cvpr06} and the Event-8
    dataset~\citep{event-8}, for varying training label noise $R$ and
    varying number of training trials $T$. Experiments on other
    datasets obtain the same trend. Ensemble learning is able to
    cancell out the deficiency of the training sets even it is very
    severe (e.g. $R=80\%$), given that the deficiency modes are
    different or `orthogonal' and the number of training sets are
    sufficiently large. The figure is best viewed in color.}
\label{fig:mov2}
\end{figure*}

\subsection{Observation 2}
\label{sec:mov2}
Ensemble learning is able to cancel out or  substantially mitigate the
deficiency of individual training sets, given that the number of such
training sets is sufficiently large and the modes of the deficiency
are different or `orthogonal'. 

We examined this idea in supervised
image categorization.  Given the ground truth data divided into training
and test sets: $\mathcal D = \{\mathcal{D}^\text{train} ,
\mathcal{D}^\text{test}\}$, (i) we artificially synthesized a set of
weak training sets (training sets with different modes of deficiency)
$\mathcal{D}^\text{train}_t, t=1,\ldots,T$ from training data
$\mathcal{D}^\text{train}$, and (ii) ensemble learning was then
performed on these sets and its performance on test data
classification was measured.

In order to guarantee the diversity of the training sets (for ensemble
learning), each weak training set $\mathcal{D}^\text{train}_t$ is
formed by randomly taking $30\%$ of the images in $\dtrain$, and
randomly re-assigning labels of a fixed percentage $R$ of these
images. Hence, $R=0$ corresponds to the `oracle' performance as
every sample is assigned its true label.  A classifier is trained for
each of these weak training sets.  At test time, each of these
classifiers returns the class label of each image in
$\mathcal{D}^\text{test}$. The winning label is the mode of the
results returned by all the classifiers.  Figure~\ref{fig:mov2}
evaluates this for the Scene-15
dataset~\citep{lazebnik:cvpr06}. Logistic Regression is used as the
classifiers with the CNN feature~\citep{deep:bmvc14} as
input.  When the label noise percentage $R$ is low, the classification
precision starts out high and levels quickly with $T$, as one would
expect. But interestingly, for $R$ even as high as $80\%$, the
classification precision, which starts low, converges to a similarly
high precision given sufficient weak training sets $T$ ($\approx
500$).  This suggests that ensemble learning is able to cancel out the
deficiency of individual training sets. It learns the essence of image
classes when the modes of deficiency are different for different
training sets, and given a sufficiently large number of such training sets.

We were inspired by the two observations, and would like to
investigate whether the assumptions of \emph{local-consis\-tency} and
\emph{exotic-consistency} are enough to generate a set of such weak
training sets in an unsupervised manner, with which ensemble learning
is able to learn useful visual attributes for semi-supervised
image classification and image clustering.


\section{Our Approach}
\label{sec:approach}


The training data consists of both labeled data $\mathcal{D}_l =
\{(\bold{x}_i, y_i)\}_{i=1}^{l}$ and unlabeled data $\mathcal{D}_u =
\{\bold{x}_i\}_{i=l+1}^{l+u}$, where $\bold{x}_i$ denotes the feature
vector of image $i$, $y_i \in\{1, ..., C\}$ represents its label, and
$C$ is the number of classes. For image clustering, $l=0$, and
$u$ is the total number of images.  Most previous semi-supervised
learning (SSL) methods learn a classifier $\phi: \mathcal{X} \mapsto
\mathcal{Y}$ from $\mathcal{D}_l$ with a regulation term learned from
$\mathcal{D}_u$. Our method learns a new image representation
$\bold{f}$ from all known data $\mathcal{D} = \mathcal{D}_l \cup
\mathcal{D}_u$, and trains standard classifier models $\phi$ with
$\bold{f}$.  $\bold{f}_i$ is a vector of similarities of image $i$ to
a series of sampled image prototypes.

Let us assume that Ensemble Projection (EP) learns knowledge from $T$
prototype sets $\mathcal{P}^{t} = \{(s_i^t,
c_i^t)\}_{i=1}^{r\times n}$, where  $t\in \{1, ..., T \}$, $s_i^t \in \{1, ..., l+u \}$ is the index
of the $i^{th}$ chosen image, $c_i^t \in \{1, ..., r\}$ is the
pseudo-label indicating which prototype $s_i^t$ belongs to.  $r$ is the
number of prototypes (surrogate classes) in
$\mathcal{P}^t$, and $n$ the number of images sampled for each
prototype (class) (\eg $r=3$ and $n=3$ in Figure~\ref{fig:pipeline}).
Below, we first present our sampling method for creating a single
prototype set $\mathcal{P}^t$ in the $t$th trial, followed by EP.

\subsection{Max-Min Sampling} 
\label{sec:max-min}
As stated, we want the prototypes to be inter-distinct and
intra-compact, so that each one represents a different visual
concept. To this end, we design a 2-step sampling method, termed
Max-Min Sampling. The Max step is based on the
\emph{exotic-consistency} and caters for the inter-distinct
property; the Min-step is based on the \emph{local-consistency}
assumption and caters for the intra-compact requirement. In particular, we first
sample a skeleton of the prototype set,  by looking for image
candidates that are strongly spread out, i.e. at large distances from
each other. We then enrich the skeleton to a prototype set by
including the closest neighbors of the skeleton images.
The algorithm for creating $\mathcal{P}^t$ is given in
Algorithm\ref{alg:max-min}. For the skeleton, we sampled $m$
hypotheses -- each one consists of $r$ randomly sampled images. 
For each hypothesis, the average pairwise distance between the $r$ images is then computed. 
Finally, we take the hypothesis yielding the largest average mutual distance as the skeleton. 
This simple
procedure guarantees that the sampled seed images are far from each
other. Once the skeleton is created, the Min-step extends each seed
image to an image prototype by introducing its $n$ nearest neighbors
(including itself), in order to enrich the characteristics of each
image prototype and reduce the risk of introducing noisy images. The
pseudo-labels are shared by all images specifying the same prototype.
It is worth pointing out that the randomized Max-step may not generate
the optimal skeleton. However, it serves its purpose well. For one thing, we do not need the optimal one -- we only need the prototypes
to be \emph{far apart}, not \emph{farthest apart}.  Moreover, randomization allows diverse visual
concepts to be captured in different $\mathcal{P}^t$'s. 
The influence of the optimality of each single skeleton is tested in Section~\ref{sec:parameters}.
The Euclidean distance is used here, but it is easy to change to other distance metrics if needed. 

\subsection{Ensemble Projection}
We now explore the use of the image prototype sets created in
Section~\ref{sec:max-min} for a new image representation.  Because the
prototypes are compact in feature space, each of them implicitly
defines a visual concept. This is especially true
when the dataset $\mathcal{D}$ is large, which is to be
expected given the vast number of unlabeled images that are
available.  Since the information carried by a single prototype set
$\mathcal{P}^t$ is quite limited and noisy, we borrow an idea from
ensemble learning (EL), namely to create an ensemble of $T$ such sets to
accumulate wisdom from a brood set of training images. A sanity check of this was
already presented for a simulated situation in Section~\ref{sec:mov2}.

As is well-known~\citep{zhou:ensemble}, EL benefits from the precision of its base learners
and their diversity.  
To obtain a good precision, discriminative learning method is employed for the
base learner $\phi_t(.)$; logistic regression is used in our
implementation to project each input image $\vect{x}$ to the image
prototypes to measure the similarities. This choice is both due to its
training efficiency and because lower capacity models are better suited
for the sparse, small-size datasets under consideration.  To achieve a high
diversity, randomness is introduced in different trials of Max-Min
Sampling to create an ensemble of diverse prototype sets, so that a
rich set of image attributes are captured.
The vector of all similarities is then concatenated and used as a new
image representation $\vect{f}$ for the final classification. A
standard classifier (\eg SVMs, Boosting, or Random Forest) can then be
trained on $\mathcal{D}_l$ with the learned feature $\vect{f}$ for the
semi-supervised classification, as unlabeled data has already been
explored when obtaining $\vect{f}$. Likely, image clustering is
performed by injecting the learned feature to a standard clustering
method.  The whole procedure of EP is presented in
Algorithm\ref{alg:ensemble:projection}.  By now, the whole pipeline in
Figure\ref{fig:pipeline} has been explained.

\begin{algorithm}[!t]\small
  \KwData{$ \text{Dataset } \mathcal{D}$}
\KwResult{$ \text{Prototype set } \mathcal{P}^t$}
\Begin{ $ \hat{e} = 0$ \tcc*[r]{Max-step} \While{$\text{iterations}
    \leq m$}{ $\mathcal{V} = \{r \text{ random image indexes} \}$\; $e
    = \sum_{i\in \mathcal{V}} \sum_{j\in \mathcal{V}}
    \text{dis}(\bold{x}_i, \bold{x}_j) $\; \If{$\ e > \hat{e}$} {
      $\hat{e} = e$ \; $\hat{\mathcal{V}} = \mathcal{V}$\; } }
  \For(\tcc*[f]{Min-step}){$i \gets 1$ \KwTo $r$ } { $\vect{s}_i^t =
    \text{stacked indexes of the } n \text{ nearest neighbors of }
    \mathcal{V}(i) \text{ in } \mathcal{D}$\; 
   $\vect{c}_i^t = (i, i,...,i) \in \mathbb{R}^n$\;
  } 
$\vect{s}^t = (\vect{s}_1^t, ..., \vect{s}_r^t) \in \mathbb{R}^{r\times n}$  
  $\vect{c}^t = (\vect{c}_1^t, ..., \vect{c}_r^t) \in   \mathbb{R}^{r\times n} $\;
 $\mathcal{P}^t = \{(s_i^t, c_i^t)\}_{i=1}^{r\times n}$  \;
}
\caption{Max-Min Sampling in $t^{th}$ trial}
\label{alg:max-min}
\end{algorithm}

 


\begin{algorithm}[!t] \small
  \KwData{$\text{Dataset } \mathcal{D} \text{ with image presentation } {\vect{x}_i}$}
  \KwResult{$\text{Dataset } \mathcal{D} \text{ with image presentation } {\vect{f}_i}$}
\Begin{
 \For{$t \gets 1$ \KwTo $T$}{
$\text{Sample } \mathcal{P}^t = \{(s_i^t, c_i^t)\}_{i=1}^{r\times n} \text{using Algorithm~\ref{alg:max-min}}$ \;
$\text{Train classifiers }\phi^t(.) \in \{1, ..., r\} \text{ on } \mathcal{P}^t$ \; 
 }

 \For{$i \gets 1$ \KwTo $l+u$}{
 \For{$t \gets 1$ \KwTo $T$}{
$\text{Obtain projection vector: } \vect{f}^t_i = \phi^t({\vect{x}_i})$ \; 
}
$\vect{f}_i = ((\vect{f}_i^1)^\top, ..., (\vect{f}_i^T)^\top)^\top $ \; 
}
}
\caption{Ensemble Projection } 
\label{alg:ensemble:projection}
\end{algorithm}

\section{Experiments}
\label{sec:experiments}

The effectiveness of the approach is evaluated in the situations of:
(1) semi-supervised image classification, where the amount of labeled
data is sparse relative to the total amount of data; and (2) image
clustering, where no labeled data is provided. In this section, we
will first introduce the datasets and the features used, followed
by experimental results for the two tasks and their corresponding analysis.

\textbf{Datasets:} The method is evaluated on diverse classification
tasks: texture classification, object classification, scene
classification, event classification, style classification, and
satellite image classification. Eight standard datasets are used for
the evaluation:

\begin{itemize}
\item Texture-25~\citep{UIUC:Texture}: $25$
texture classes, with $40$ samples per class.

\item Caltech-101~\citep{FeiFei2004}: $101$ object classes, with
  $31$ to $800$ images per class, and $8677$ images in total,

\item STL-10~\citep{stl-10}: $10$ object classes including airplane, bird,
  car, cat, deer, dog, horse, monkey, ship, truck, with $500$ training
  images per class, $800$ test images per class, and $100000$
  unlabeled images for unsupervised learning.

\item Scene-15~\citep{lazebnik:cvpr06}: $15$ scene classes with both
  indoor and outdoor environments, $4485$ images in total. Each
  class has $200$ to $400$ images.

\item Indoor-67~\citep{Indoor}: $67$ indoor classes such as shoe
  shop, mall and garage, with a total of $15620$ images and at least
  $100$ images per class.

\item Event-8~\citep{event-8}: $8$ sports event classes including rowing,
  badminton, polo, bocce, snowboarding, croquet, sailing, and rock
  climbing, with a total of $1574$ images.

\item Building-25~\citep{building-25}: $25$ architectural styles such
  as American craftsman, Baroque, and Gothic, with $4794$ images in
  total.

\item LandUse-21~\citep{landuse21}: $21$ classes of satellite images in
  terms of land usage, such as agricultural, airplane, forest. There
  are $2100$ images in total, with $100$ images per class.

\end{itemize}








\textbf{Features:} The following three features were used in our
earlier papers~\citep{dai:eccv12b, dai:iccv13b} due to their simplicity and low dimensionality: GIST
\citep{gist}, Pyramid of Histogram of Oriented Gradients (PHOG)
\citep{Bosch:iccv07}, and Local Binary Patterns (LBP)
\citep{Ojala02}. However, these features are obsolete and yield
 results inferior than alternative features recently developed  for image
classification. 
In this paper, we replaced them with the CNN features \citep{caffe14, deep:bmvc14}. 
These were obtained from an off-the-shelf CNN pre-trained on the ImageNet data. 
They were chosen as CNN features
have achieved state-of-the-art performance for image classification
\citep{nips12:cnn, cnnfet14}. For implementation, we used the
MatConvNet \citep{MatConvNet} toolbox, with a $21$-layer CNN
pre-trained model being used. The convolutional results at layer $16$ were
stacked as the CNN feature vector, with dimensionality of $4096$.  We
also tested the LLE-coded SIFT feature \citep{siftllc:cvpr10}. However,
it is not on par with the CNN features.
 


\textbf{Competing methods:} For semi-supervised classification, six
classifiers were adopted to evaluate the method, with three baselines:
$k$-NN, Logistic Regression (LR), and SVMs with RBF kernels, and three
semi-supervised classifiers: Harmonic Function (HF)
\citep{Zhu:Harmonic:03}, LapSVM \citep{Belkin:semiframe:2006}, and
Anchor Graph (AG) \citep{icml10:large:graph:ssl}. HF formulates the SSL
learning problem as a Gaussian Random Field on a graph for label
propagation. LapSVM extends SVMs by including a smoothness penalty
term defined on the Laplacian graph. AG aims to address the
scalability issue of graph-based SSL, and constructs a tractable large
graph by coupling anchor-based label prediction and adjacency matrix
design.  For image clustering, we compare our learned feature to the
original CNN feature with two standard clustering algorithms: $k$-means
and Spectral Clustering. Existing systems for image clustering often
report performance on relatively easy datasets and it is hard to compare
with them on these standard classification datasets.


\textbf{Experimental settings:} We conducted four sets of experiments:
(1) compare our method with competing methods for semi-supervised
image classification on the eight datasets, where the unlabeled images
are from the same class as the labeled ones; (2) evaluate the
robustness of our method against the choice of its parameters and classifier models
in the context of semi-supervised image classification; (3) evaluate
the performance of our method for the task of self-taught image
classification on the STL-10 dataset, where the feature is learned from
the unlabeled images and the performance is tested on the labeled set;
and (4) evaluate our method for the task of image clustering on the
eight datasets.

For all experimental setups except (2), the same set of parameters were
used for all the classifiers. We used $k=1$ for the $k$-NN classifier,
L2-regularized LR of LIBLINEAR~\citep{liblinear} with $C=15$, and the
SVMs with RBF kernel of LIBSVM~\citep{libsvm} with $C=15$ and the
default $g$, \ie $g=1/4096$. For LapSVM, we used the scheme suggested
by~\citep{Belkin:semiframe:2006}: $\gamma_A$ was set as the inductive
model, and $\gamma_I$ was set as $\frac{\gamma_Il}{(l+u)^2} =
100\gamma_Al$. For HF, the weight matrix was computed with the
Gaussian function $e^{-|| \vect{x}_i - \vect{x}_j ||^2 / 2 \sigma^2}$,
where $\sigma$ is automatically set by using the self-tuning
method~\citep{selftuning:04}. For AG, we followed the suggestion from
the original work~\citep{icml10:large:graph:ssl} and used the following
for both our learned feature and the original CNN feature: $1000$
anchors and features reduced to $500$ dimensions via PCA.

As to the parameters of our method, a wide variety of values for them
were tested in experimental setup (2). In experimental setups (1), (3) and
(4), we fixed them to the following values: $T=100$, $r=30$, $n=6$,
and $m=50$, which leads to a feature vector of $3000$ dimensions. Note
that the learned feature may contain redundancy across different
dimensions, as some prototype sets are similar to others. We leave the
task of selecting useful features to the discriminative classifiers.
 


\begin{figure*} 
  \centering
   $ \begin{array}{cccc} \hspace{-1mm}
\includegraphics[width=0.24\linewidth, height=32mm]{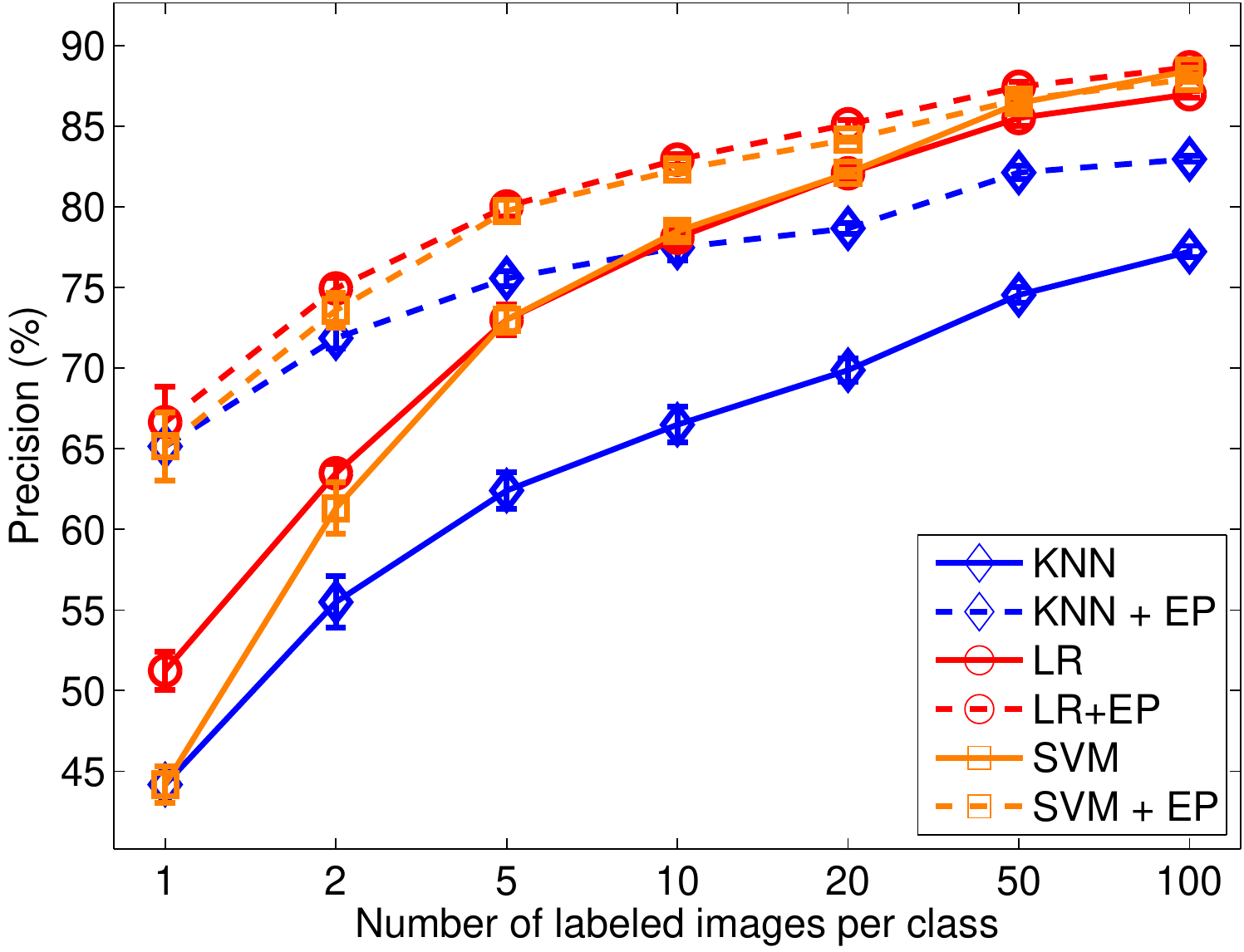}& 
\hspace{-2mm}
\includegraphics[width=0.24\linewidth, height=32mm]{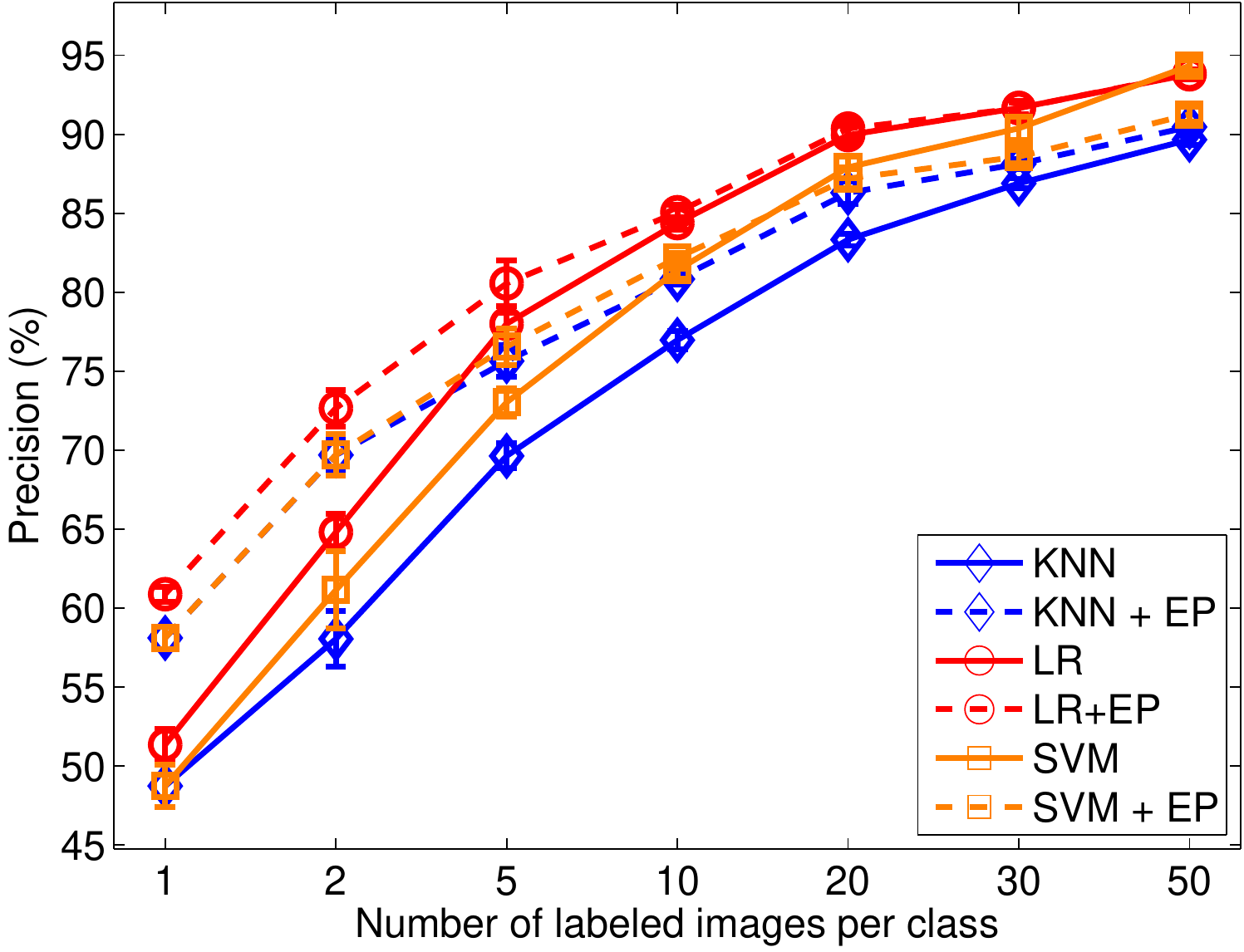}& \hspace{-2mm}
\includegraphics[width=0.24\linewidth, height=32mm]{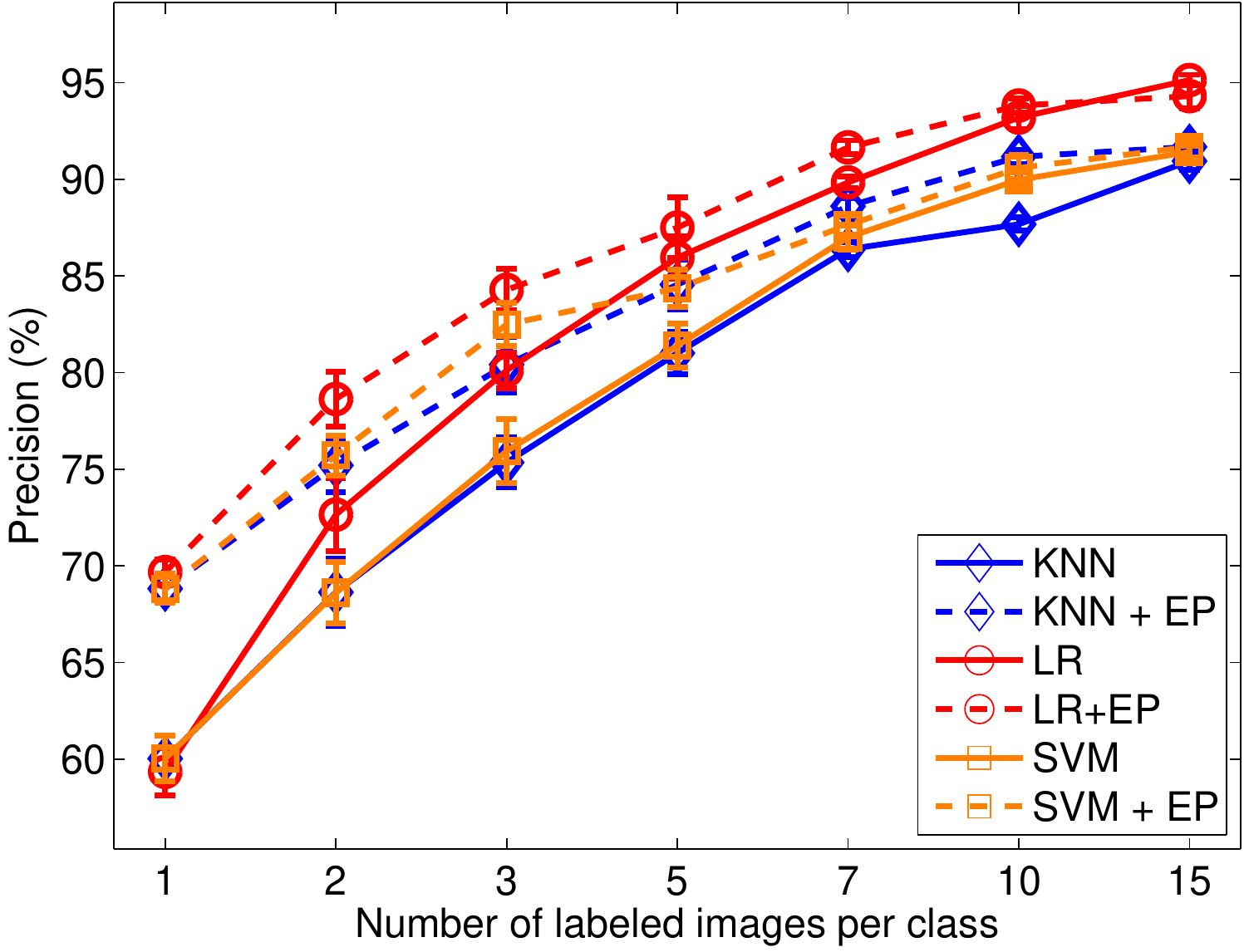}&  \hspace{-2mm}
\includegraphics[width=0.24\linewidth, height=32mm]{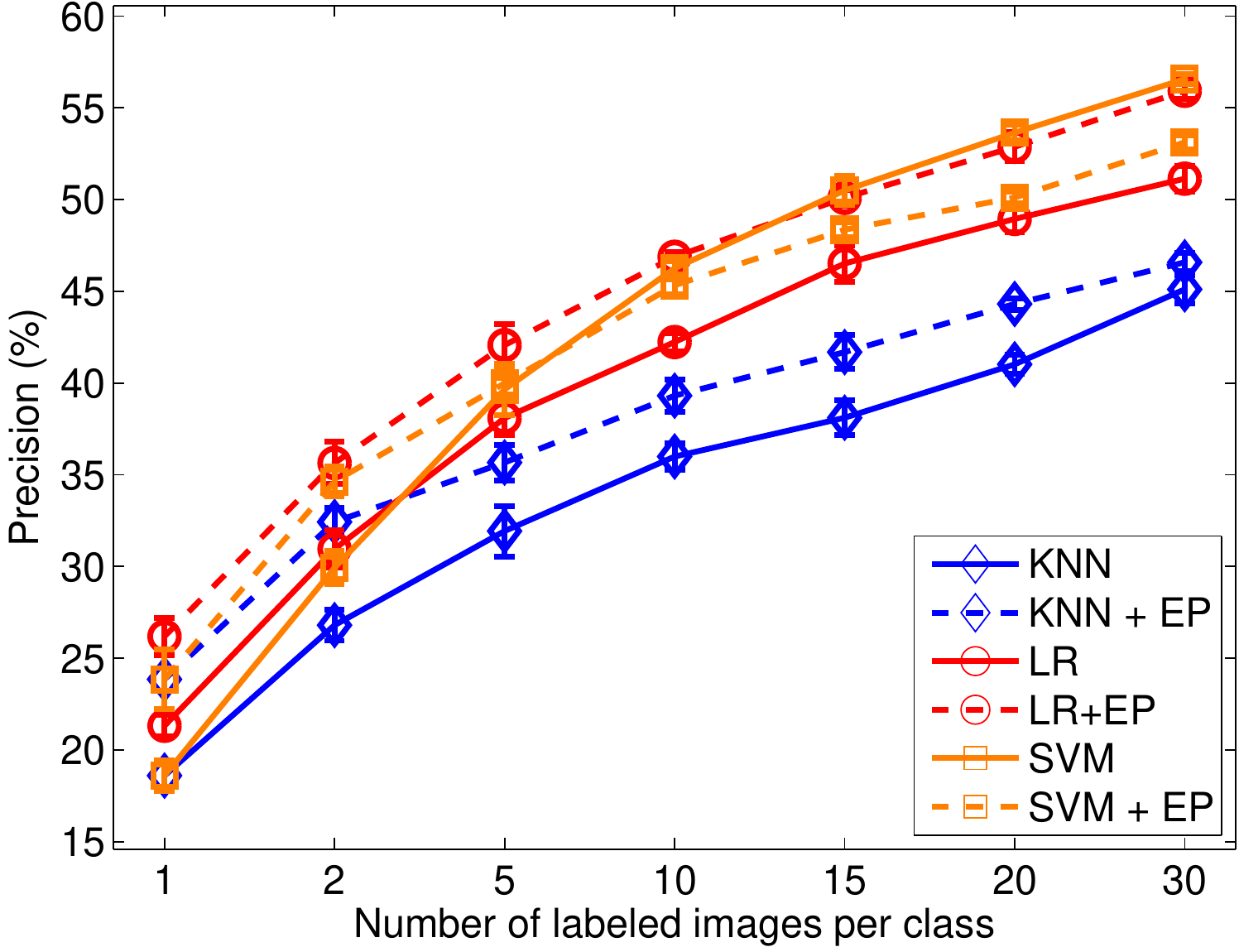}   \\
\scriptsize{\text{(a) Scene-15}} & \scriptsize{\text{(b) LandUse-21}} & \scriptsize{\text{(c) Texture-25}} & \scriptsize{\text{(d) Building-25}}  \\

\hspace{-1mm}
\includegraphics[width=0.24\linewidth, height=32mm]{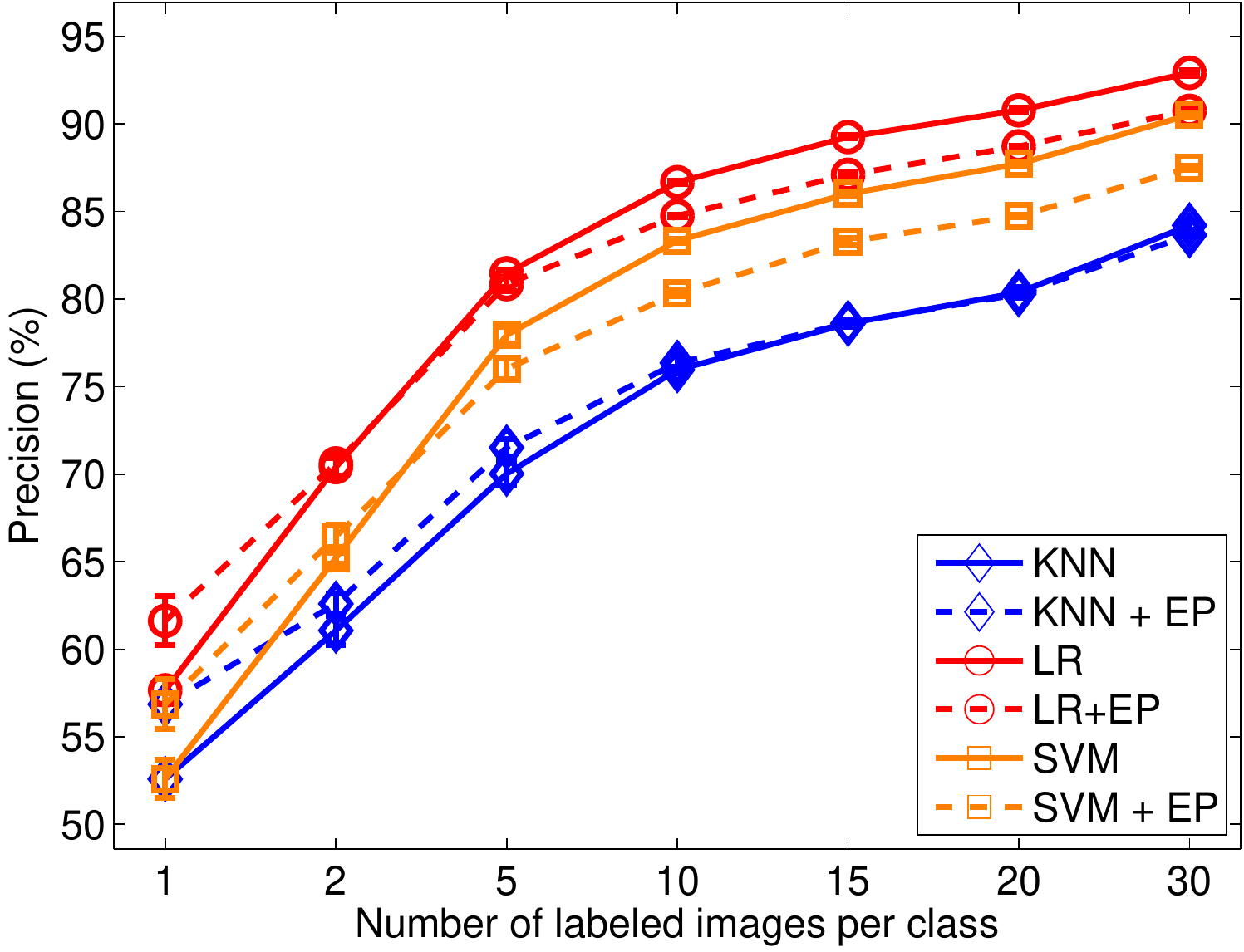}& \hspace{-2mm}
\includegraphics[width=0.24\linewidth, height=32mm]{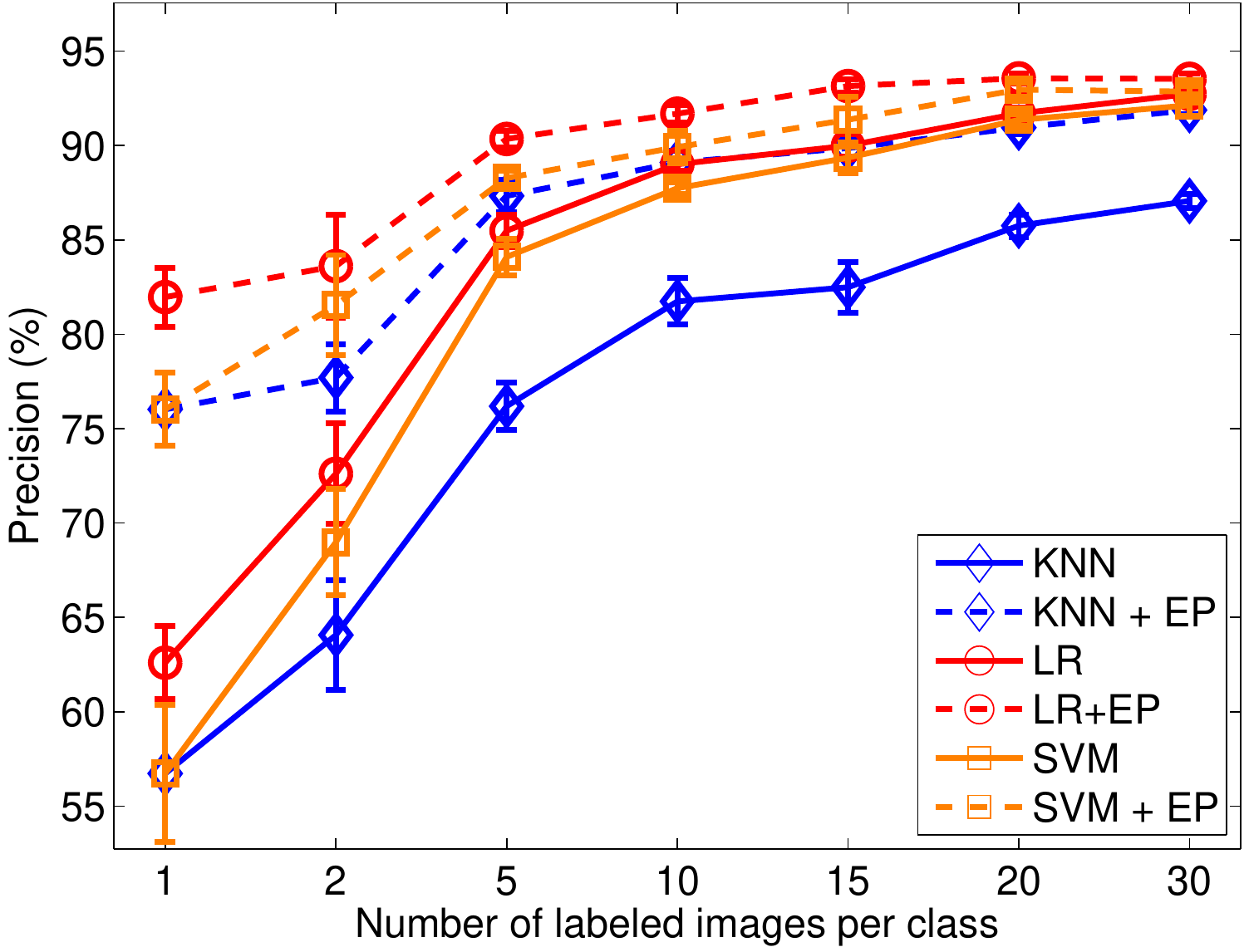}& \hspace{-2mm}
\includegraphics[width=0.24\linewidth, height=32mm]{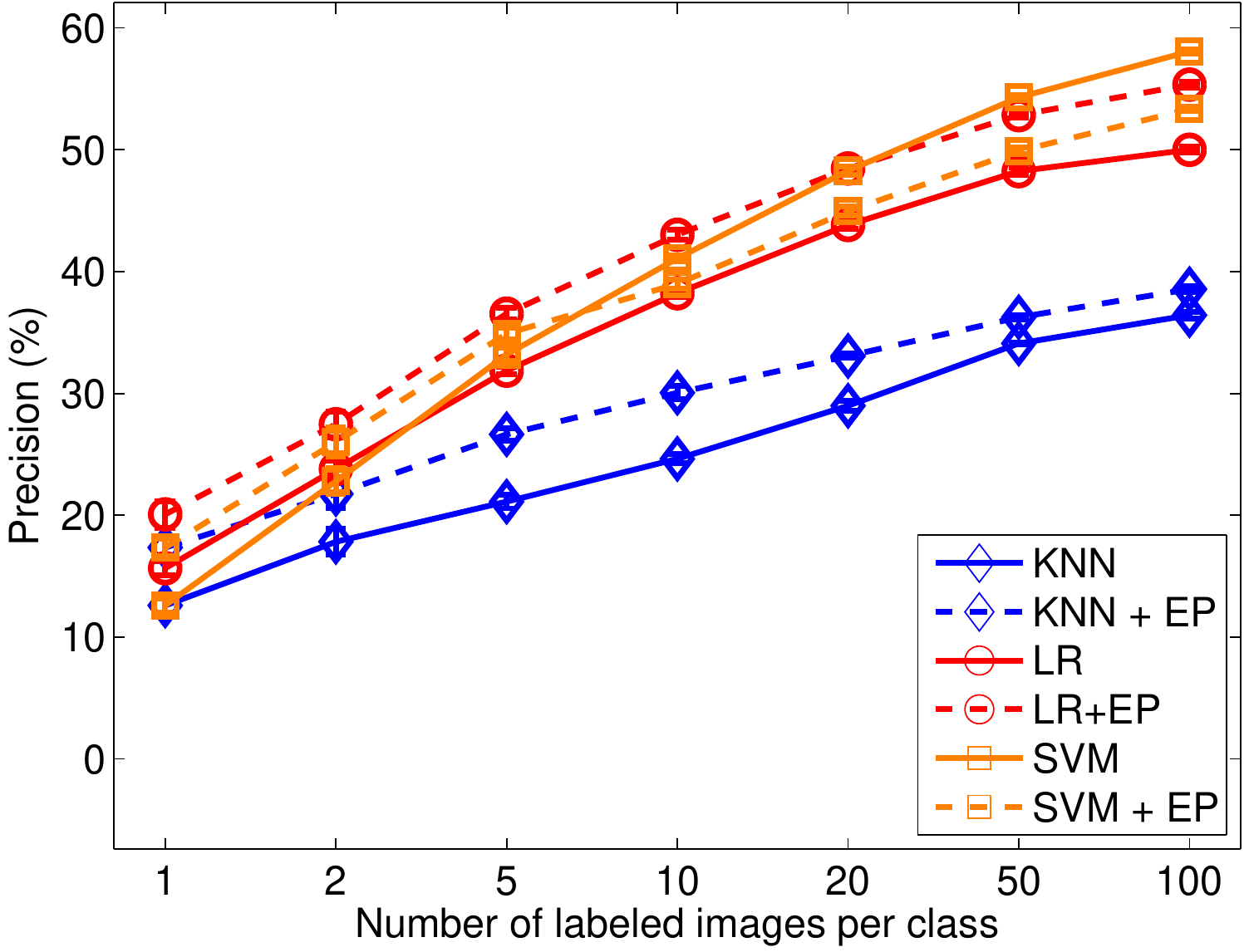}&  \hspace{-2mm}
\includegraphics[width=0.24\linewidth, height=32mm]{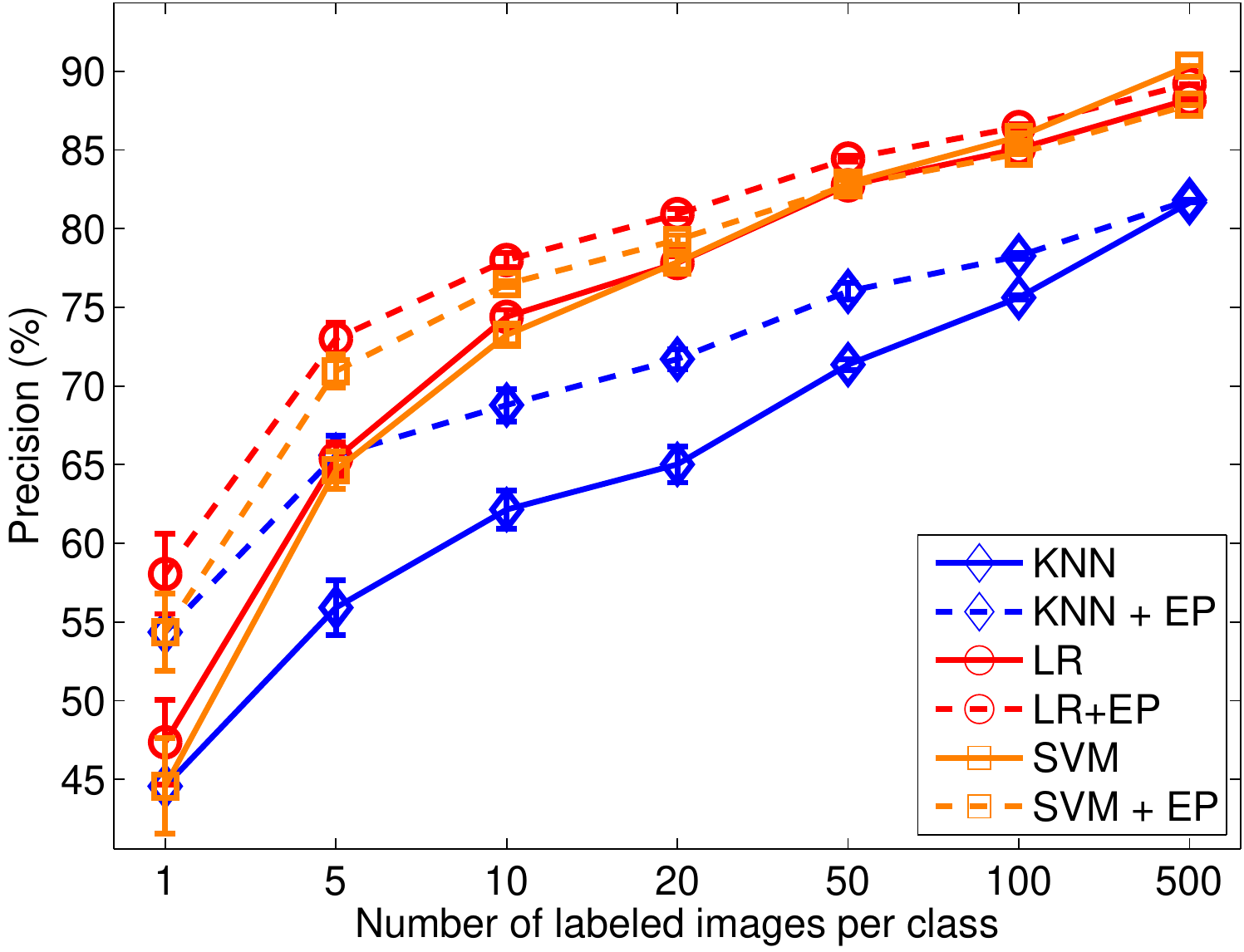}  \\ 
\scriptsize{\text{(e) Caltech-101}} & \scriptsize{\text{ (f) Event-8}} & \scriptsize{\text{(g) Indoor-67}} & \scriptsize{\text{(h) STL-10}} \\

\end{array}$
 \caption{Classification results of Ensemble Projection (EP) on the
  eight datasets, where three classifiers are used: $k$-NN, Logistic
  Regression, and SVMs with RBF kernels. All methods were tested
  with two feature inputs: the original deep feature and the learned feature by EP on top of it (indicated by ``+ EP"). }
  \label{fig:results:baseline}
\end{figure*}

\begin{figure*} 
  \centering
   $ \begin{array}{cccc}
\hspace{-1mm}
\includegraphics[width=0.24\linewidth, height=32mm]{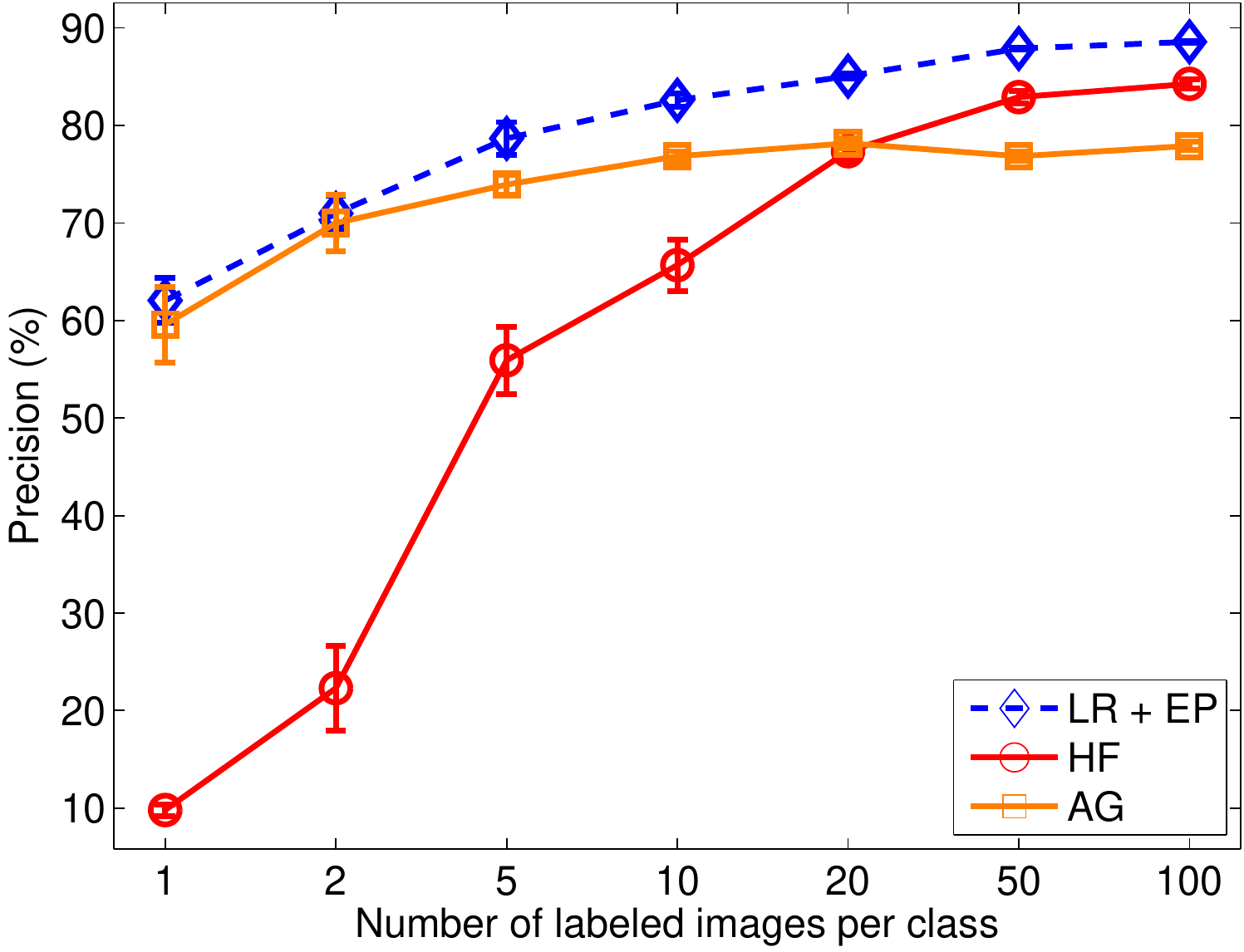}& \hspace{-2mm}
\includegraphics[width=0.24\linewidth, height=32mm]{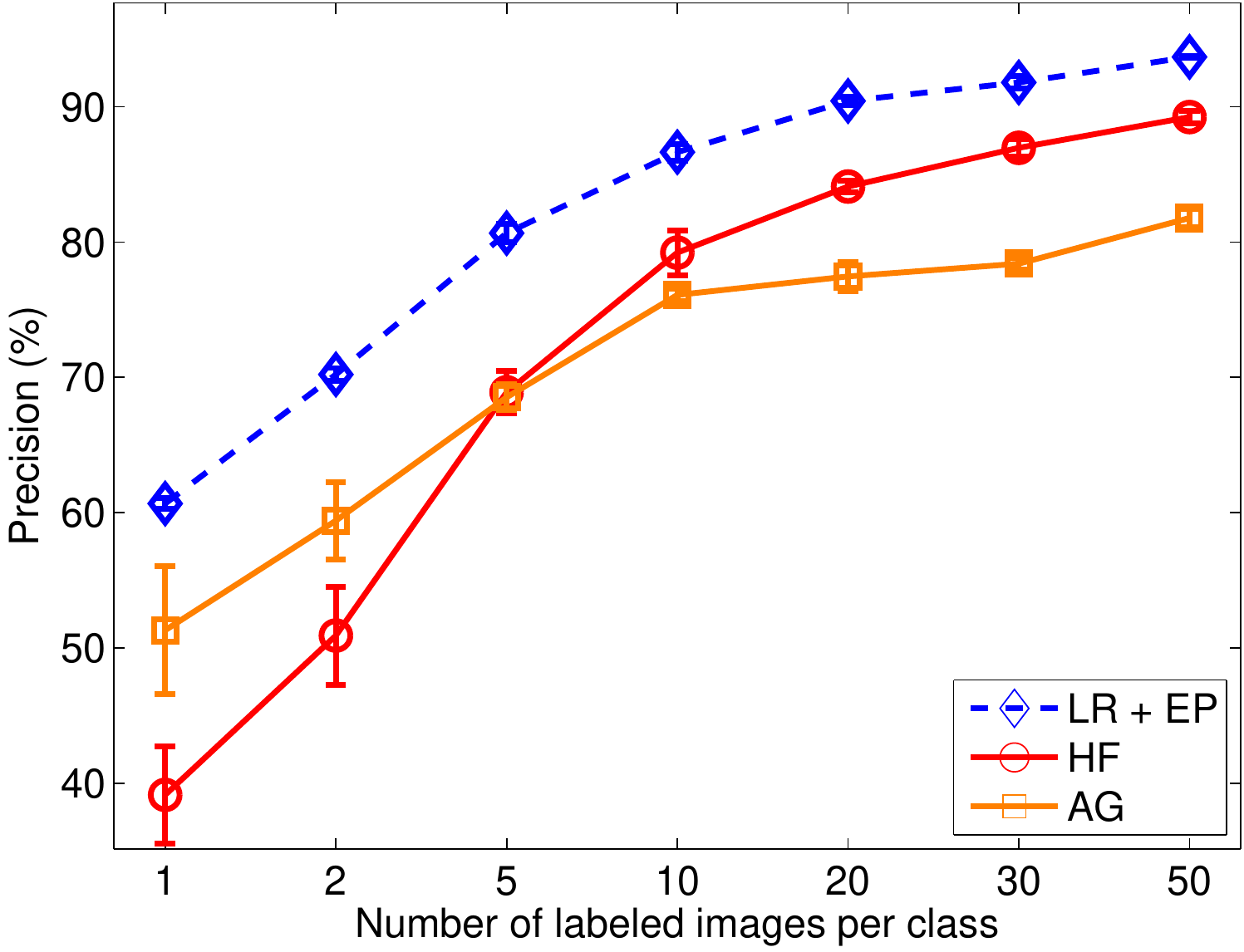}& \hspace{-2mm}
\includegraphics[width=0.24\linewidth, height=32mm]{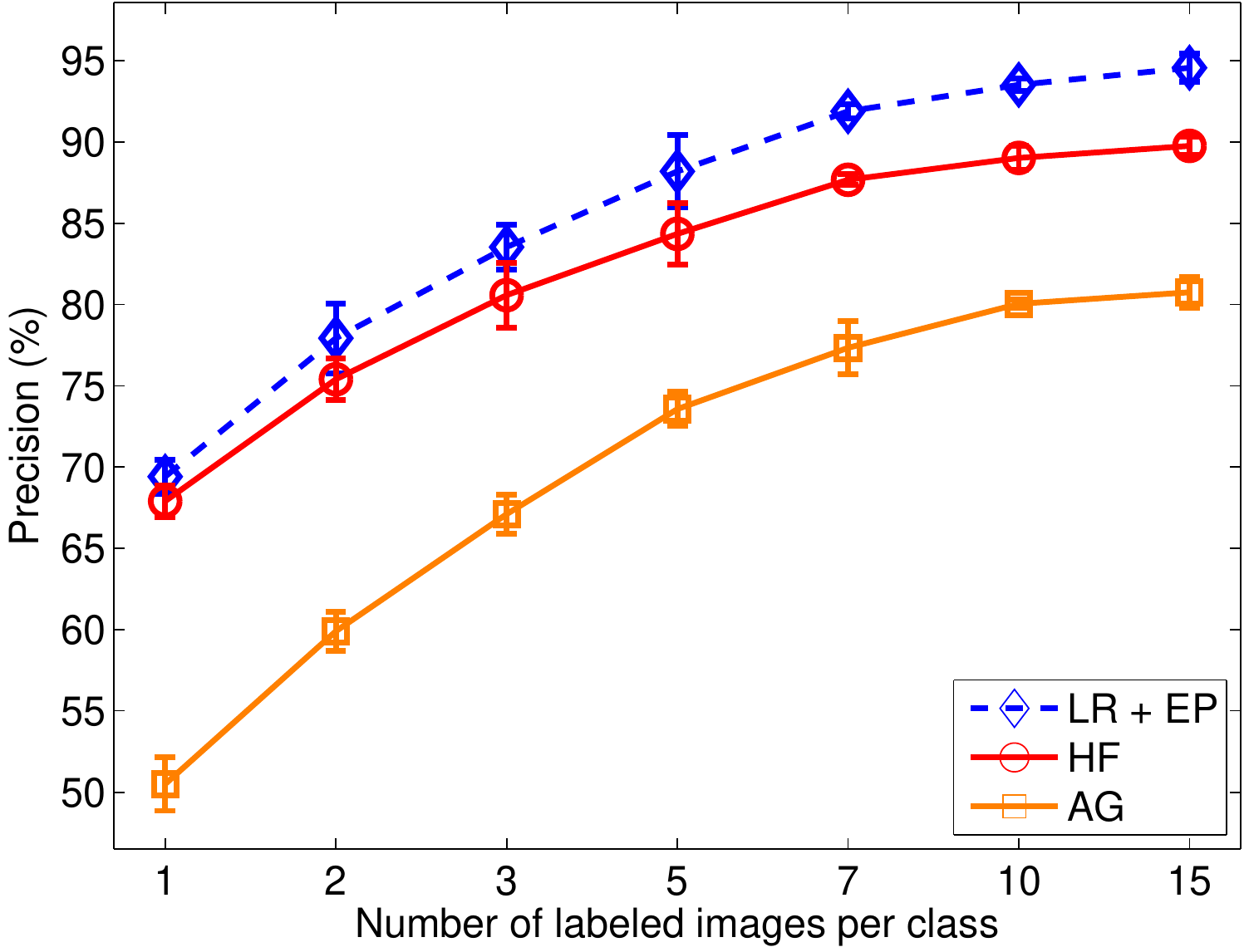}& \hspace{-2mm}
\includegraphics[width=0.24\linewidth, height=32mm]{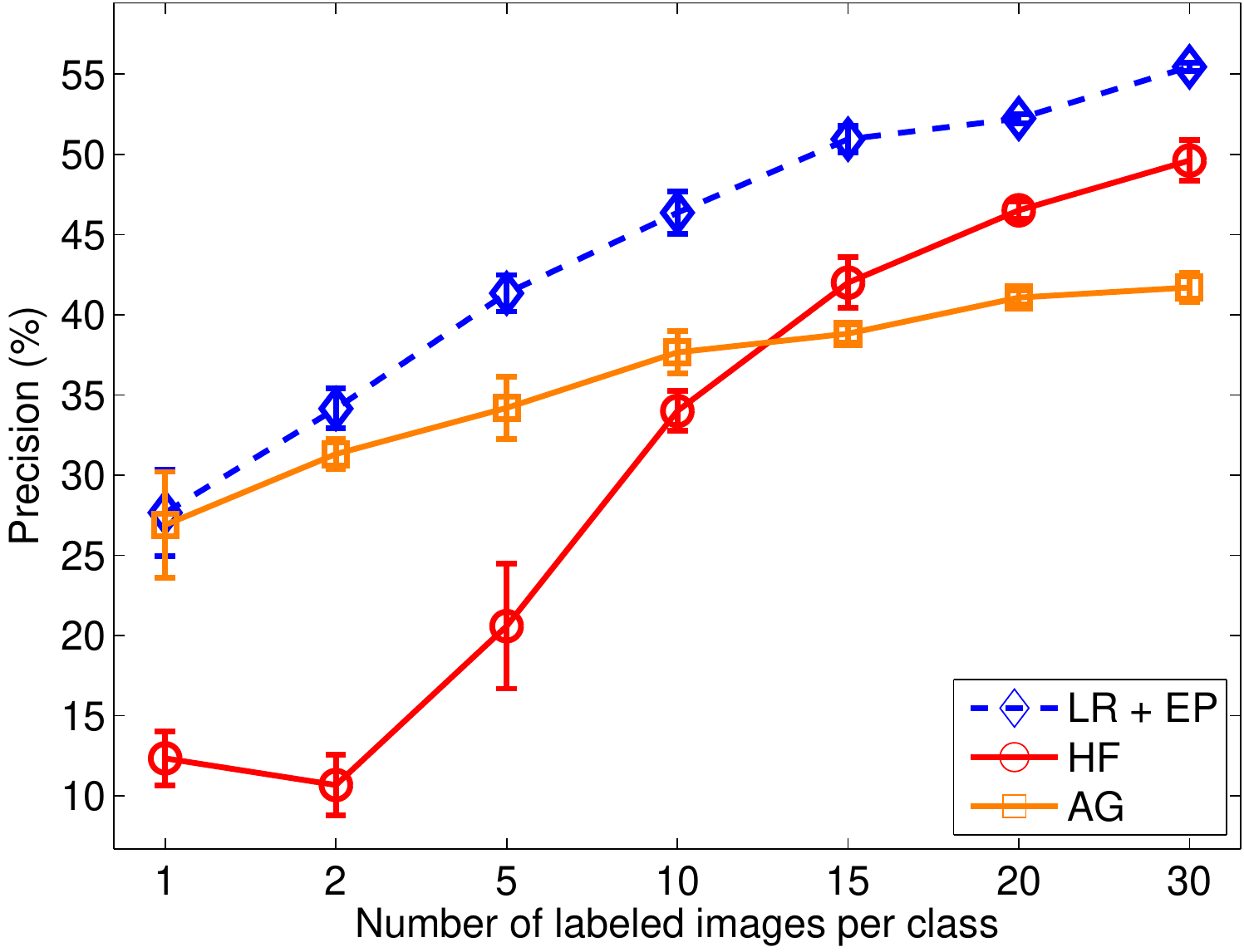}   \\
\scriptsize{\text{(a) Scene-15}} & \scriptsize{\text{ (b) LandUse-21}} & \scriptsize{\text{(c) Texture-25}} & \scriptsize{\text{(d) Building-25}}  \\

\hspace{-1mm}
\includegraphics[width=0.24\linewidth, height=32mm]{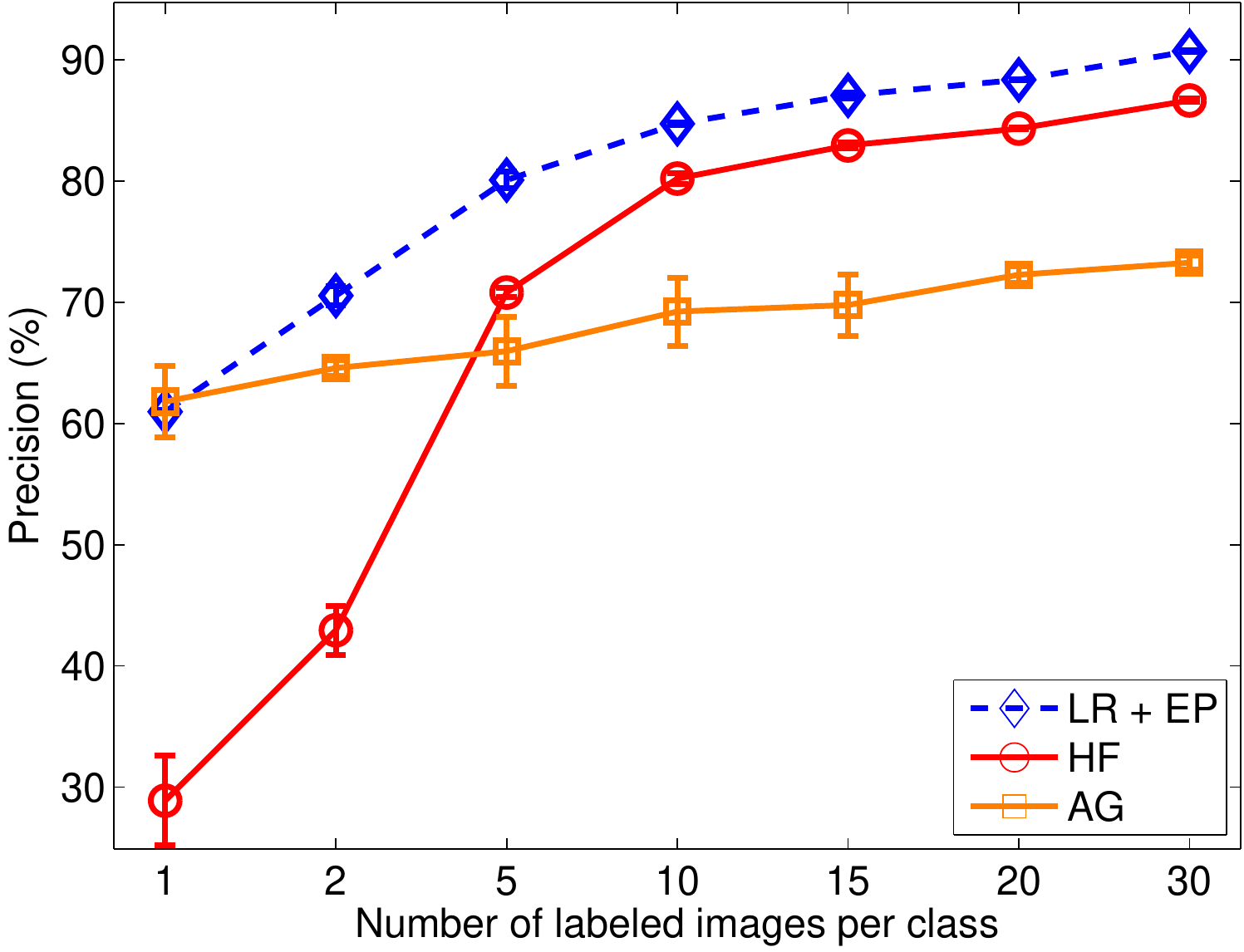}& \hspace{-1mm}
\includegraphics[width=0.24\linewidth, height=32mm]{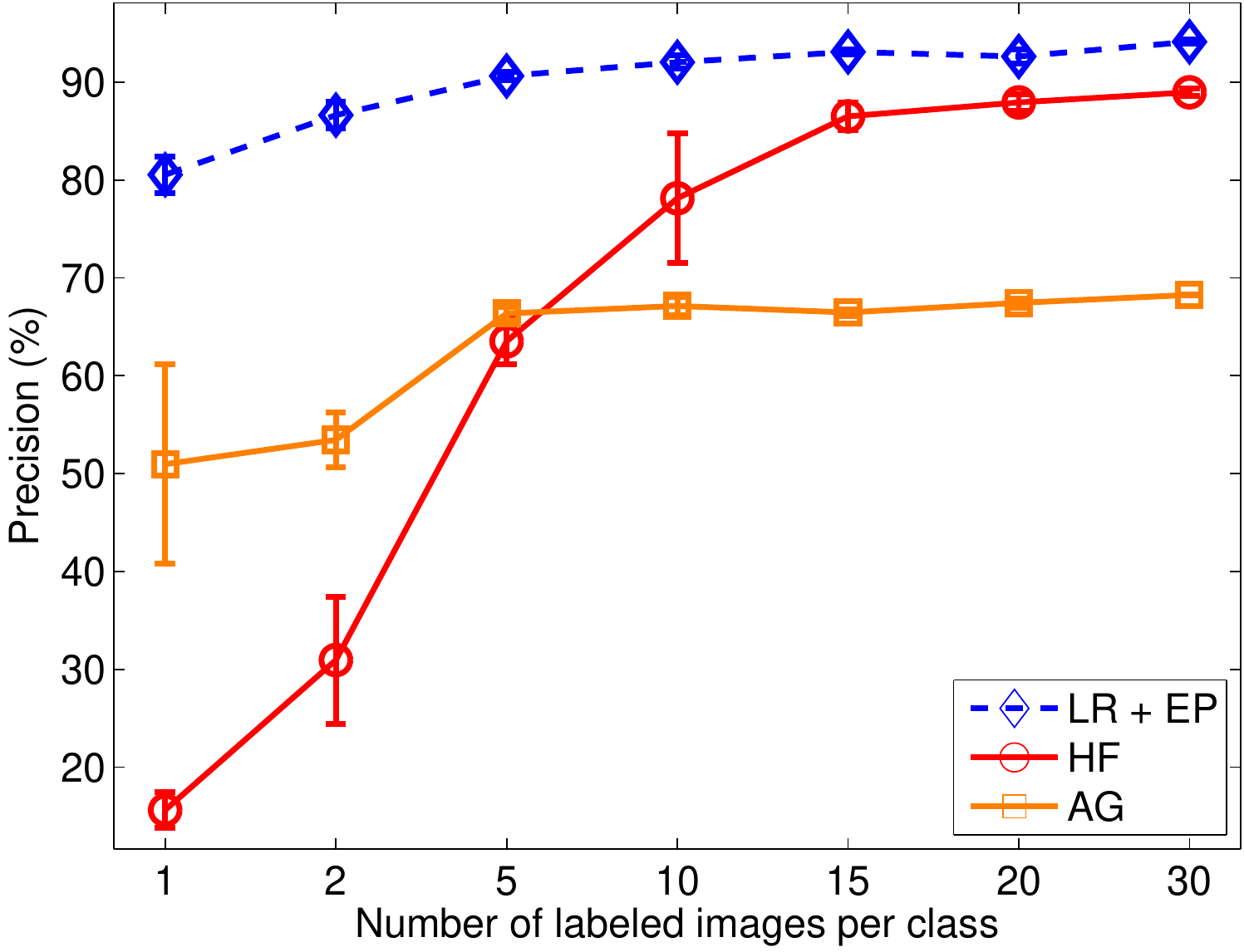}& \hspace{-2mm}
\includegraphics[width=0.24\linewidth, height=32mm]{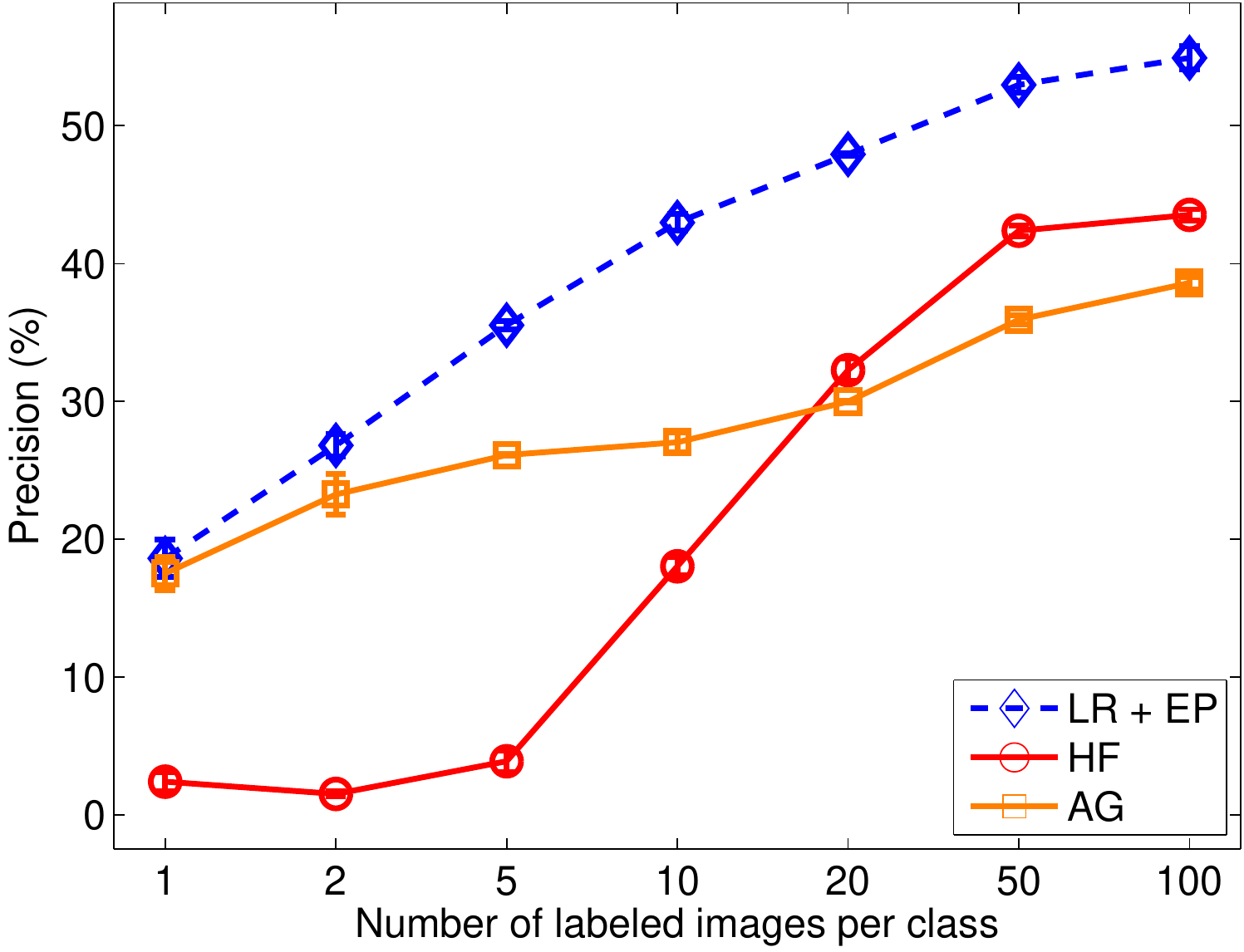}& \hspace{-2mm}
\includegraphics[width=0.24\linewidth, height=32mm]{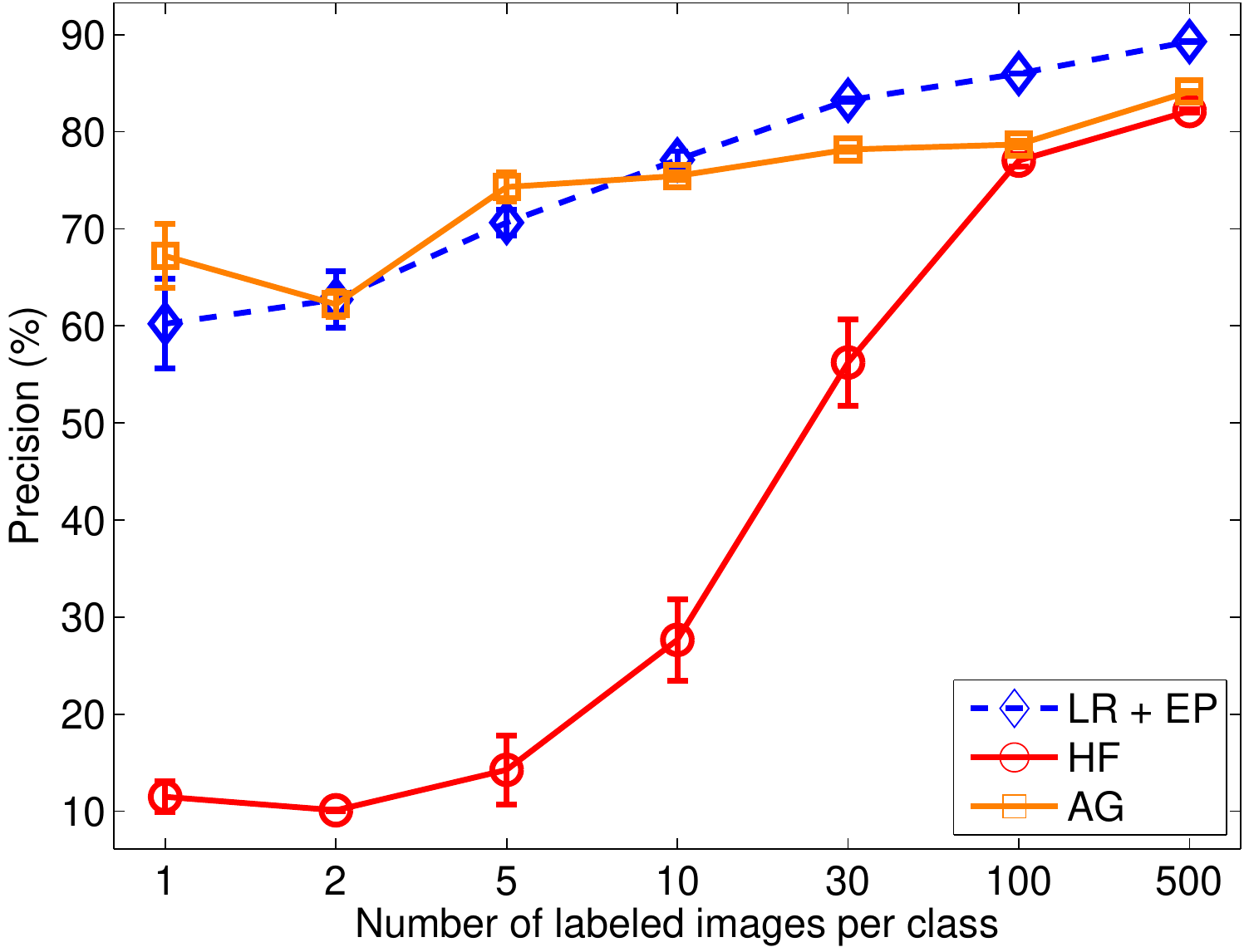}  \\ 
\scriptsize{\text{(e) Caltech-101}} & \scriptsize{\text{(f) Event-8}} & \scriptsize{\text{(g) Indoor-67}} & \scriptsize{\text{(h) STL-10}} \\

\end{array}$
\caption{Classification results of Ensemble Projection (EP) on the
  eight datasets, where three classifiers are used: $k$-NN, Logistic
  Regression, and SVMs with RBF kernels. All methods were tested
  with two feature inputs: the original deep feature and the learned feature by EP on top of it (indicated by ``+ EP"). }
  \label{fig:results:ssl}
\end{figure*}


\begin{table*}[tb] \small 
\setlength\tabcolsep{0.20em} {
$\begin{array}{ccccc}
\begin{tabular}{cc}
\toprule 
\multicolumn{1}{c}{\footnotesize{Methods}} \\ \midrule 
\emph{k}-NN   \\
\emph{k}-NN + \bf{EP}  \\
LR     \\
LR + \bf{EP} \\  
SVMs     \\
SVMs + \bf{EP} \\  
HF     \\
HF + \bf{EP} \\  
AG     \\
AG + \bf{EP} \\  
\bottomrule
\end{tabular}

\begin{tabular}{cc}
\toprule
\multicolumn{1}{c}{\footnotesize{Scene-15}} \\ \midrule
62.4 (1.4) \\
75.6 (0.6) \\
73.0 (1.2) \\
\textbf{80.0} (0.8) \\
73.0 (1.0) \\
\underline{79.8} (0.8) \\
45.2 (0.3) \\
61.5 (0.2) \\
72.8 (1.1) \\
78.3 (1.1) \\
\bottomrule
\end{tabular}

\begin{tabular}{cc}
\toprule
\multicolumn{1}{c}{\footnotesize{LandUse-21}} \\ \midrule
69.6 (1.0) \\
75.6 (1.2) \\
\underline{78.0} (0.8) \\
\textbf{80.6} (1.7) \\
73.0 (1.1) \\
76.6 (1.4) \\
39.1 (3.6) \\
52.3 (1.6) \\
51.3 (4.7) \\
58.5 (0.1) \\
\bottomrule
\end{tabular}

\begin{tabular}{cc}
\toprule
\multicolumn{1}{c}{\footnotesize{Texture-25}} \\ \midrule
81.0 (1.3) \\
\underline{84.5} (1.6) \\
85.9 (1.3) \\
\textbf{87.5} (1.9) \\
81.4 (1.4) \\
84.4 (1.2) \\
67.9 (1.0) \\
74.6 (1.0) \\
50.5 (1.6) \\
32.1 (2.5) \\
\bottomrule
\end{tabular}

\begin{tabular}{cc}
\toprule
\multicolumn{1}{c}{\footnotesize{Building-25}} \\ \midrule
31.9 (1.7) \\
35.7 (1.2) \\
38.1 (1.1) \\
\textbf{42.1} (1.4) \\
39.6 (1.7) \\
\underline{40.0} (0.9) \\
18.5 (3.6) \\
21.2 (3.0) \\
34.8 (1.0) \\
37.5 (0.0) \\
\bottomrule 
\end{tabular}

\begin{tabular}{cc}
\toprule
\multicolumn{1}{c}{\footnotesize{Event-8}} \\ \midrule
76.2 (1.5) \\
87.3 (1.0) \\
85.5 (1.0) \\
\textbf{90.4} (0.5) \\
84.1 (1.2) \\
\underline{88.3} (0.4) \\
15.6 (1.8) \\
27.1 (7.7) \\
51.0 (10.2) \\
50.5 (5.6) \\
\bottomrule
\end{tabular}

\begin{tabular}{cc}
\toprule
\multicolumn{1}{c}{\footnotesize{Caltech-101}} \\ \midrule
70.0 (0.8) \\
71.5 (0.6) \\
\textbf{81.5} (0.2) \\
\underline{80.9} (0.5) \\
77.9 (0.5) \\
76.0 (0.7) \\
70.6 (0.3) \\
75.8 (0.4) \\
67.7 (1.4) \\
66.3 (2.2) \\
\bottomrule 
\end{tabular}

\begin{tabular}{cc}
\toprule
\multicolumn{1}{c}{\footnotesize{Indoor-67}} \\ \midrule
21.1 (0.7) \\
26.6 (0.6) \\
31.9 (0.4) \\
\textbf{36.6} (0.6) \\
33.2 (0.6) \\
\underline{34.9} (0.5) \\
7.4 (2.9) \\
12.1 (0.3) \\
24.7 (0.4) \\
24.9 (1.5) \\
\bottomrule
\end{tabular}

\begin{tabular}{cc}
\toprule
\multicolumn{1}{c}{\footnotesize{STL-10}} \\ \midrule
55.9 (2.1) \\
65.6 (1.5) \\
65.4 (1.3) \\
73.0 (1.2) \\
64.6 (1.4) \\
70.9 (1.3) \\
10.3 (0.1) \\
38.1 (8.2) \\
\textbf{76.0} (0.2) \\
\underline{74.9} (1.6) \\
 \bottomrule
\end{tabular}

\end{array} $}
\vspace{1mm}
\centering 
\caption{ Precision (\%) of image classification on the eight datasets, with $5$ labeled training examples per class. ``+ EP" indicate that classifiers working with our learned feature as input rather than the original CNN.  
  The best performance is indicated in \textbf{bold}, and the second best is \underline{underlined}. 
} \label{table:precision}
\end{table*}

\subsection{Semi-supervised Image Classification}
\label{sec:sic}
In this section, we evaluate all methods across all datasets for
semi-supervised image classification. Different numbers of training
images per class were tested: Scene-15 and Indoor-67 with \{1, 2,
5, 10, 20, 50, 100\}, LandUse-21 with \{1, 2, 5,10, 20, 30, 50\},
Texture-25 with \{1, 2, 3, 5, 7, 10 , 15\}, Building-25, Event-8,
and Caltech-101 with \{1, 2, 5, 10, 15, 20, 30\}, and STL-10
with \{1, 5, 10, 20, 50, 100, 500\}. The different choices are due
to the different structures of the datasets: different number of
classes and different number of images per class.  In keeping with
most existing systems for semi-supervised
classification~\citep{Zhu:Harmonic:03, Zhou:nips:04,
  icml10:large:graph:ssl, Fergus09, eccv10:ssl, ecml14:ssl}, we
evaluate the method in the transductive manner, where we take the
training and test samples as a whole, and randomly choose labeled
samples from the whole dataset to learn and infer labels of other
samples whose labels are held back as the unlabeled samples.  The
reported results are the average performance over $5$ runs with random
labeled-unlabeled splits.

\textbf{Comparison to baselines}: Figure~\ref{fig:results:baseline}
shows the results of the three baseline classifiers with our learned
feature and the original CNN feature as input, and
Table~\ref{table:precision} lists the results of all methods when $5$
labeled training samples are available for each class. From the figure,
it is easy to observe that the three plain classifiers $k$-NN, LR and
SVMs perform consistently better when working with our feature than
working with the original CNN features.  This is, of course, not a fair
comparison, as our feature has been learned with the help of unlabeled
samples, while the CNN features not. However, this experiment
serves as a good sanity check: given the access to the unlabeled
samples, does the proposed feature learning improve the performance of
the system over the original feature? The figure shows clear
advantages of our method over the original CNN feature across
different datasets and classifiers. The most pronounced improvement
occurs in the scenarios where a small number of labeled training
samples is available, \eg from $1$ to $5$. This is exactly what the
method is designed for -- classification tasks where the labeled
training samples are sparse relative to the available unlabeled
samples.  Since LR performs generally the best when working with our
learned feature, we will take LR + EP as our method to compare to
other SSL methods. The comparison is made in the next section.

\textbf{Comparison to other SSL Methods}: In this section, we compare
our method (LR + EP) with the three SSL methods HF, AG, and LapSVM.
The classification precision is reported for HF and AG, while the mean average
precision (mAP) of $C$ rounds of binary classification is used for
LapSVM. This is because the implementation of LapSVM from the authors 
performs binary classification \citep{Belkin:semiframe:2006}. Because LapSVM is computationally
expensive, we only compare our method to it for the scenario where $5$
labeled training samples per class are used.

Figure~\ref{fig:results:ssl} shows the results of our method (LR + EP)
and that of HF and AG, and Table~\ref{table:precision} lists the
precision of the methods when $5$ labeled training examples per class
are used. Table~\ref{table:map} lists the mAP of our method, HF and
LapSVM, when $5$ labeled training samples are available for each
class. The figure and the tables show that our method outperforms the
competing SSL methods consistently for semi-supervised image
classification.  For instance, if $5$ labeled training examples per
class are used, our method (LR + EP) improves over the best competing
method AG by 7.2\% in terms of precision on Scene-15, and by 11.9\% on
Indoor-67. This suggests that our method can achieve superior results
for semi-supervised image classification, even when combined with very
standard classifiers. It can be found from the figure and tables that
graph-based SSL methods such as HF and AG are not very stable. This is
mainly due to their sensitivity to the graph structure, which was
observed in \citep{nips14:ssl} as well.

The superior performance of our method to other SSL methods can be ascribed to two factors: (1) in addition to the
\emph{local-consistency} assumption, our method also exploits the
\emph{exotic-consistency} assumption; (2) the discriminative
projections abstract high-level attributes from the sampled
prototypes, \eg being more ``yellow-smooth'' than
``dark-structured''.  As already proven in fully supervised
scenarios~\citep{ObjectAttribute:cvpr09, Transfer:CVPR:08},
prototype-linked, attribute-based features are very helpful for image
classification.  
The superior performance of our method to the original feature~\citep{deep:bmvc14} is that 
our method learns the statistics of the to-be-classified dataset, while standard CNN features are trained
on a different dataset, though very large. The exploitation of
 dataset-specific properties by EP can be understood as 
feature enhancing or fine-tuning in an unsupervised manner.   

We further investigate the complementarity of our learned feature with
other SSL methods for semi-supervised classification. It is
interesting to see from the bottom panel of
Table~\ref{table:precision} that using such combinations boosts the
performance also. This suggests that our scheme of exploiting
unlabeled data and the previous ones doing so capture complementary
information. However, using the standard Logistic Regression generally
yields the best results for our learned feature.





\begin{table*}[tb] \small 
\setlength\tabcolsep{0.24em} {
$\begin{array}{ccccc}
\begin{tabular}{cc}
\toprule
\multicolumn{1}{c}{\footnotesize{Methods}} \\ \midrule
LR + \bf{EP} \\  
HF  \\
LapSVM  \\ \bottomrule
\end{tabular}

\begin{tabular}{cc}
\toprule
\multicolumn{1}{c}{\footnotesize{Scene-15}} \\ \midrule
\bf{84.8} (1.3) \\
81.4 (1.9) \\
79.2 (2.2) \\ 
\bottomrule
\end{tabular}

\begin{tabular}{cc}
\toprule
\multicolumn{1}{c}{\footnotesize{LandUse-21}} \\ \midrule
\bf{85.6} (1.0) \\
84.2 (0.9) \\
82.3 (0.5) \\ \bottomrule
\end{tabular}

\begin{tabular}{cc}
\toprule
\multicolumn{1}{c}{\footnotesize{Texture-25}} \\ \midrule
\bf{95.1} (0.8) \\
94.1 (0.1) \\
91.4 (0.6) \\ \bottomrule
\end{tabular}

\begin{tabular}{cc}
\toprule
\multicolumn{1}{c}{\footnotesize{Building-25}} \\ \midrule
\bf{39.2} (1.6) \\
37.9 (1.1) \\
35.8 (1.0) \\ \bottomrule 
\end{tabular}

\begin{tabular}{cc}
\toprule
\multicolumn{1}{c}{\footnotesize{Event-8}} \\ \midrule
\bf{91.7} (0.6) \\
89.5 (0.9) \\
86.2 (0.8) \\ \bottomrule
\end{tabular}

\begin{tabular}{cc}
\toprule
\multicolumn{1}{c}{\footnotesize{Caltech-101}} \\ \midrule
\bf{73.1} (0.3) \\
71.6 (0.2) \\
56.4 (0.6) \\ \bottomrule 
\end{tabular}

\begin{tabular}{cc}
\toprule
\multicolumn{1}{c}{\footnotesize{Indoor-67}} \\ \midrule
\bf{33.2} (0.1) \\
25.1 (0.2) \\
29.3 (0.1) \\     \bottomrule
\end{tabular}

\begin{tabular}{cc}
\toprule
\multicolumn{1}{c}{\footnotesize{STL-10}} \\ \midrule
\bf{81.5} (0.8) \\
78.1 (1.0) \\
69.3 (1.2) \\ \bottomrule
\end{tabular}

\end{array} $
\vspace{1mm}
\centering 
\caption{ MAP (\%) of semi-supervised classification on the eight datasets, with $5$ labeled training examples per class. ``LR + EP" indicate Logistic Regression with our learned feature as input. 
The other two classifiers use the original CNN feature as input. The best number is indicated in \textbf{bold}. 
} \label{table:map}}
\end{table*}

\begin{figure*} 
  \centering
   $ \begin{array}{cccc}
\hspace{-1mm} 
\includegraphics[width=0.24\linewidth, height=36mm]{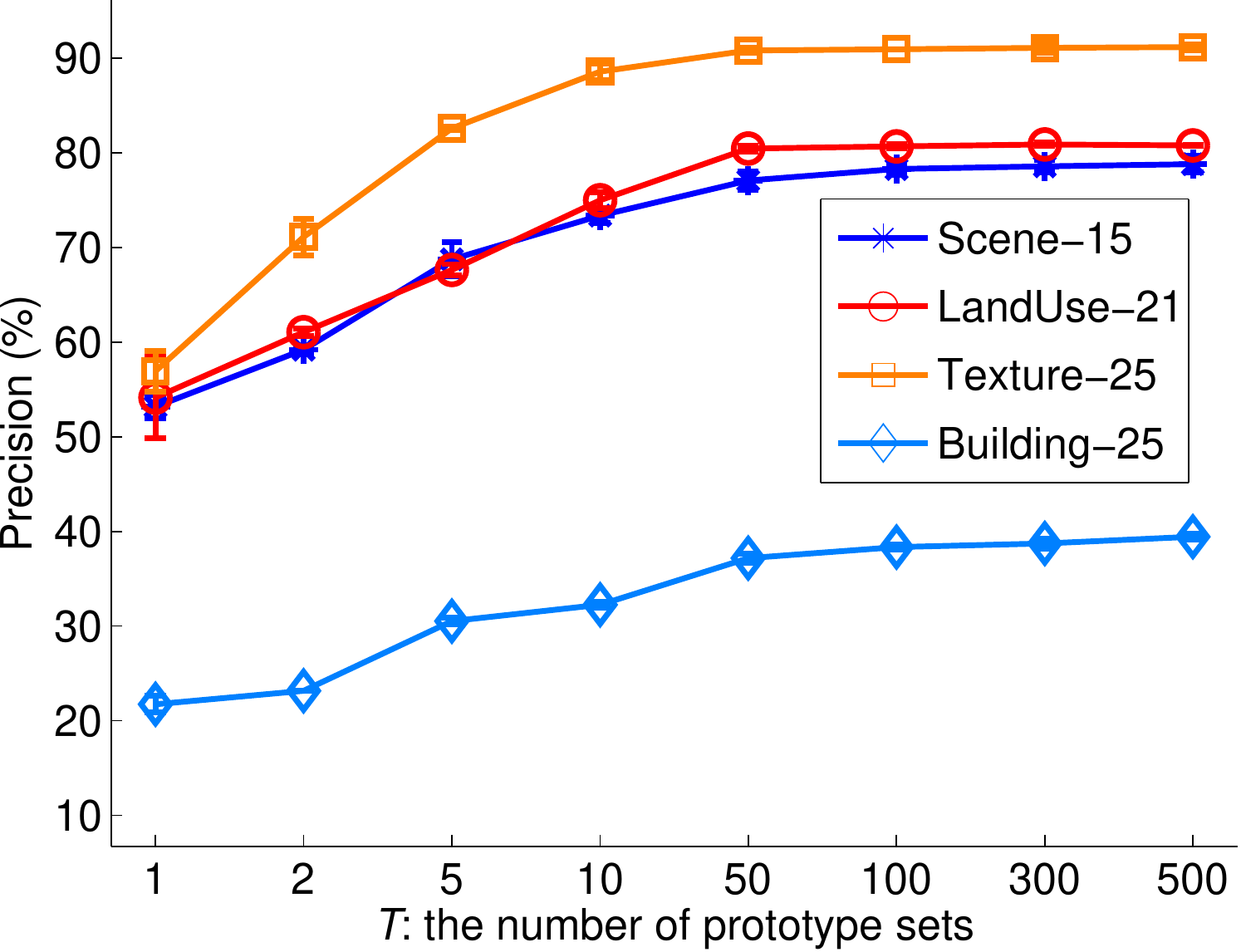} & \hspace{-2mm}
\includegraphics[width=0.24\linewidth, height=36mm]{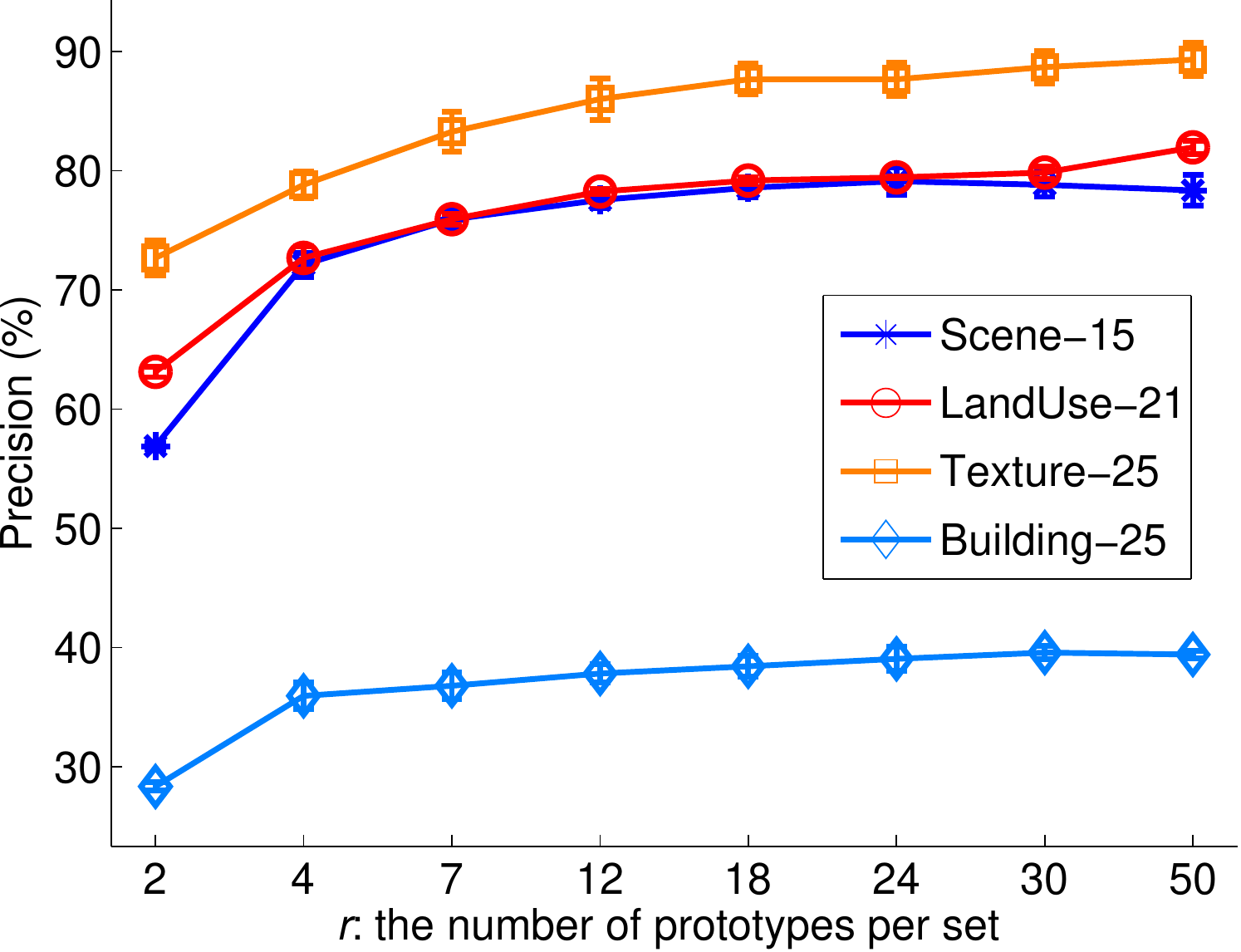} & \hspace{-2mm}
\includegraphics[width=0.24\linewidth, height=36mm]{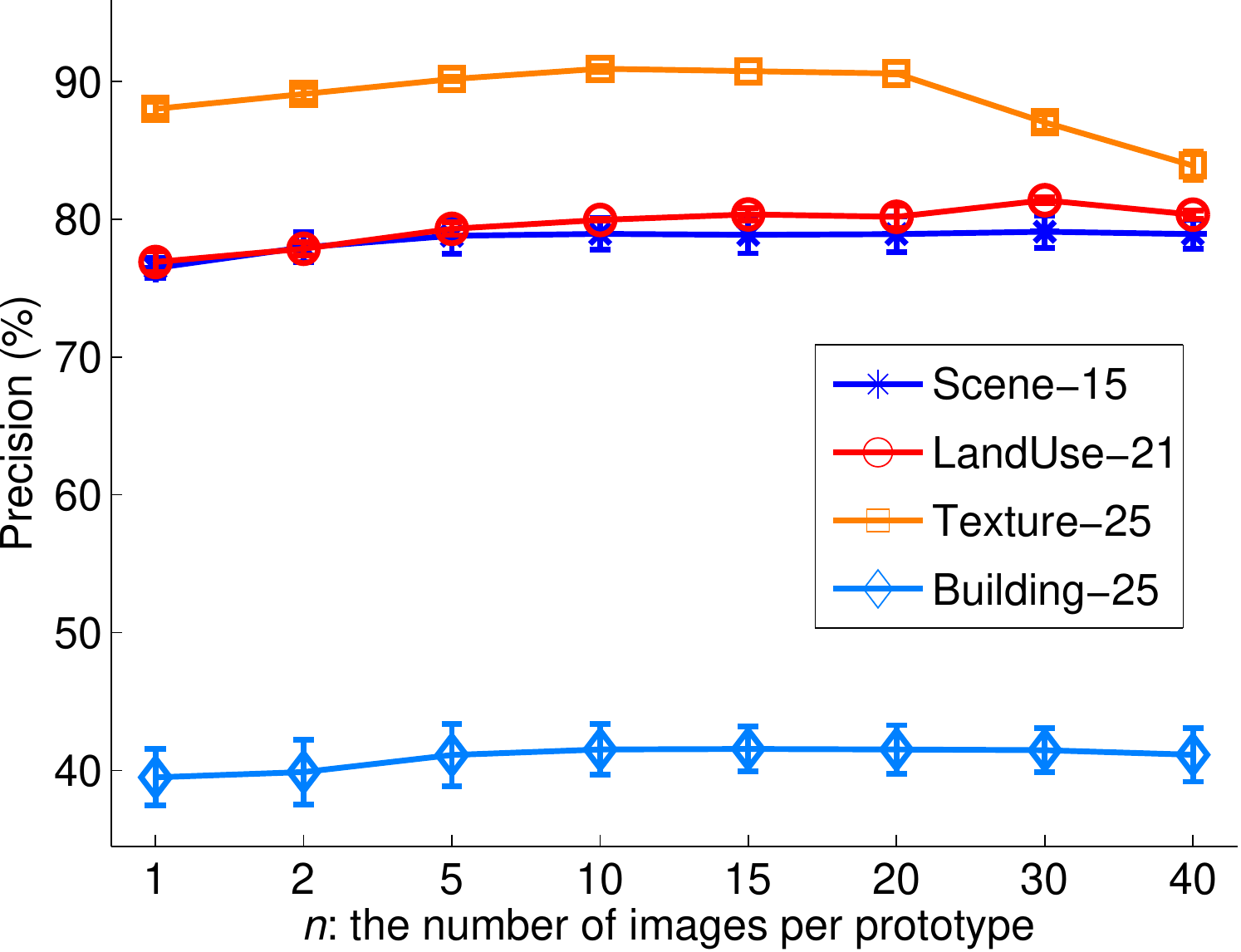} & \hspace{-2mm}
\includegraphics[width=0.24\linewidth, height=36mm]{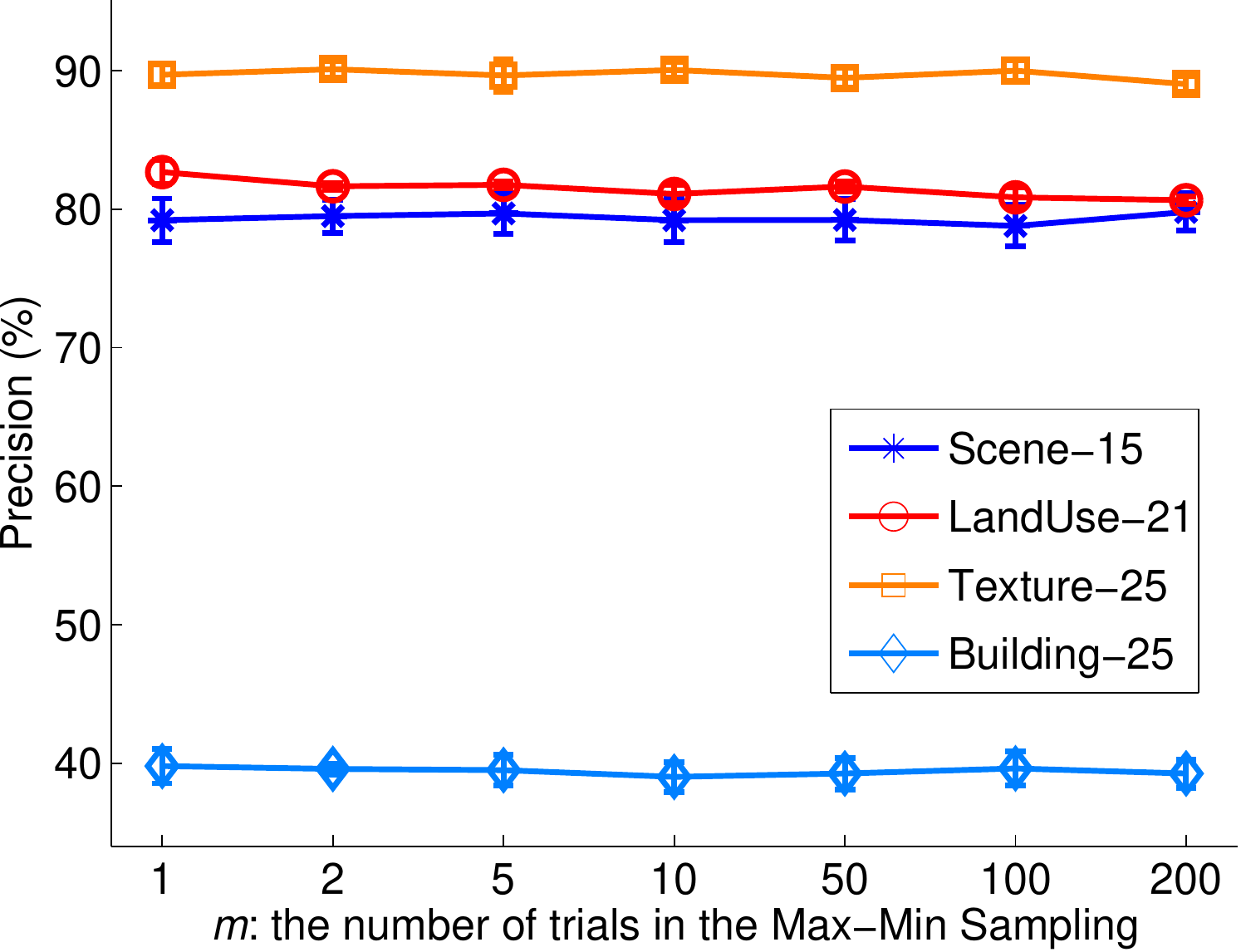} \\
\hspace{-1mm} 
\includegraphics[width=0.24\linewidth, height=36mm]{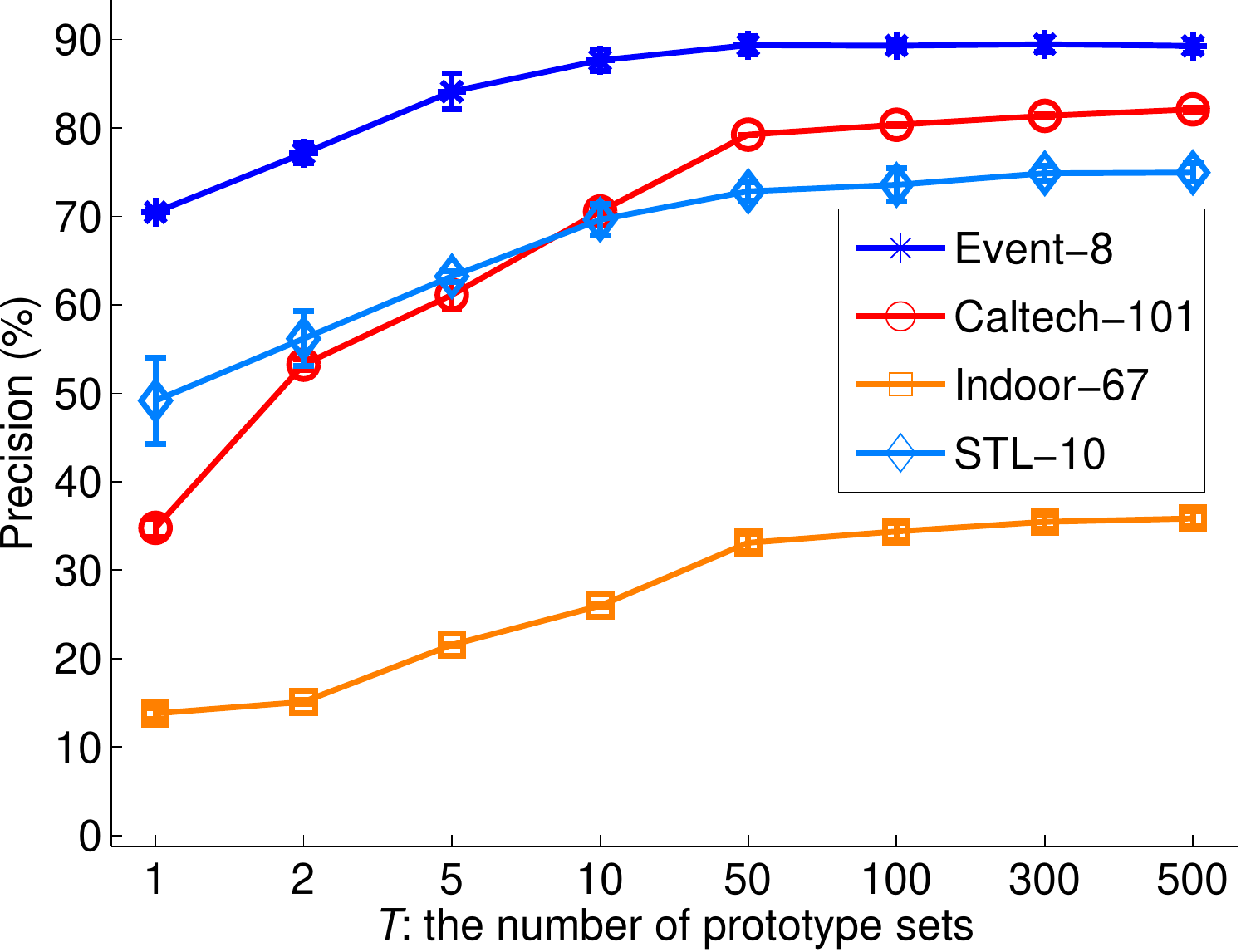}&  \hspace{-2mm}
\includegraphics[width=0.24\linewidth, height=36mm]{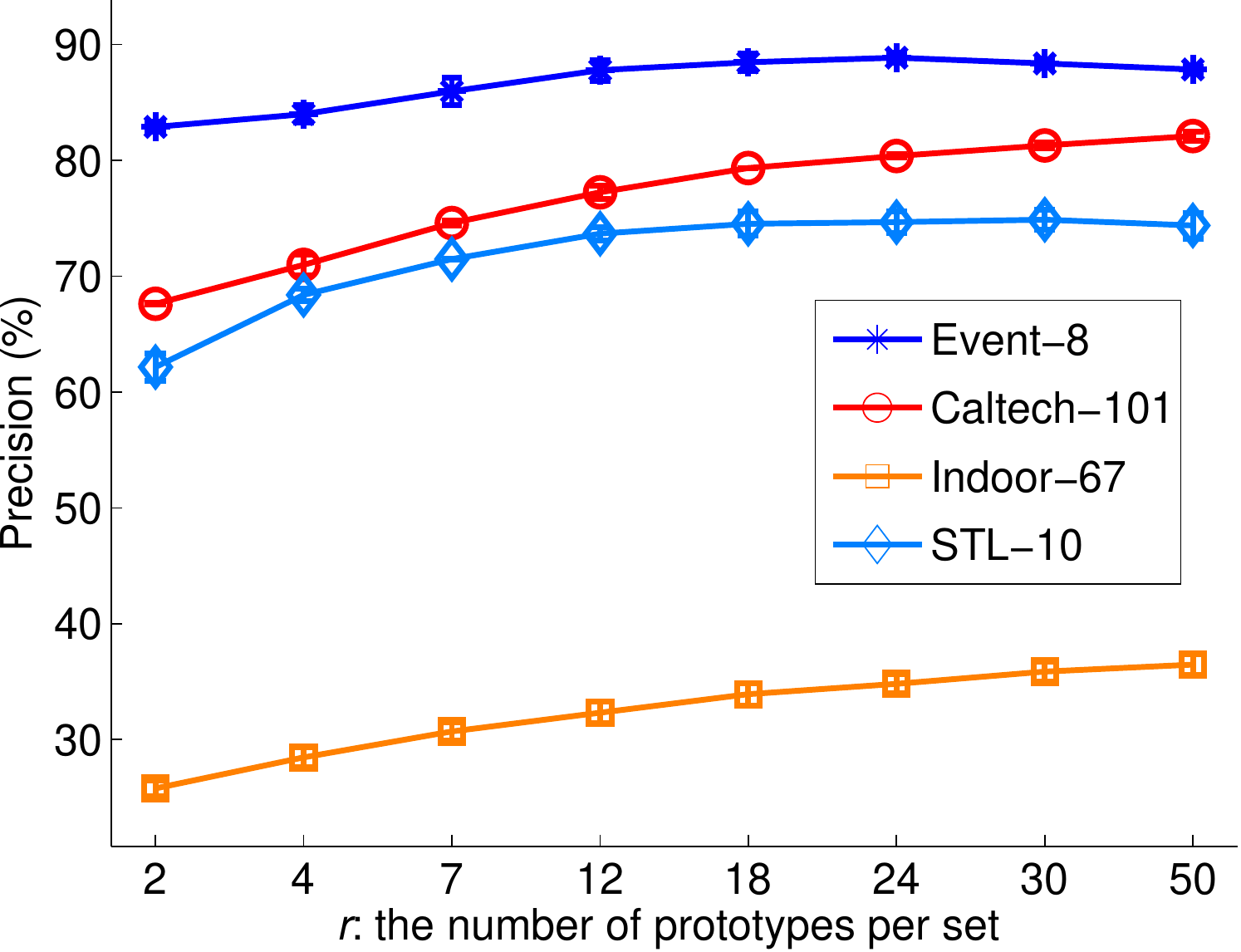} & \hspace{-2mm}
\includegraphics[width=0.24\linewidth, height=36mm]{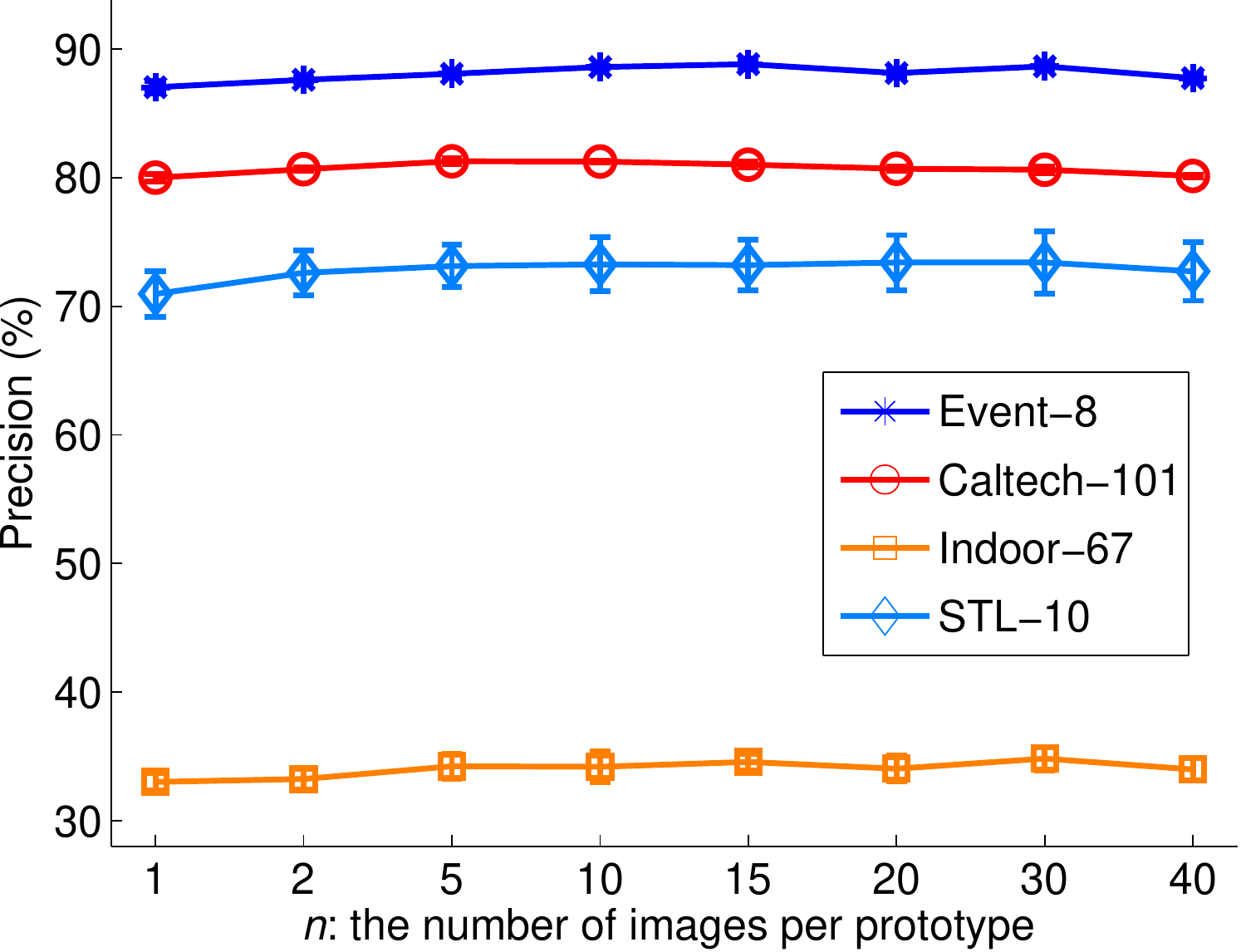}&  \hspace{-2mm}
\includegraphics[width=0.24\linewidth, height=36mm]{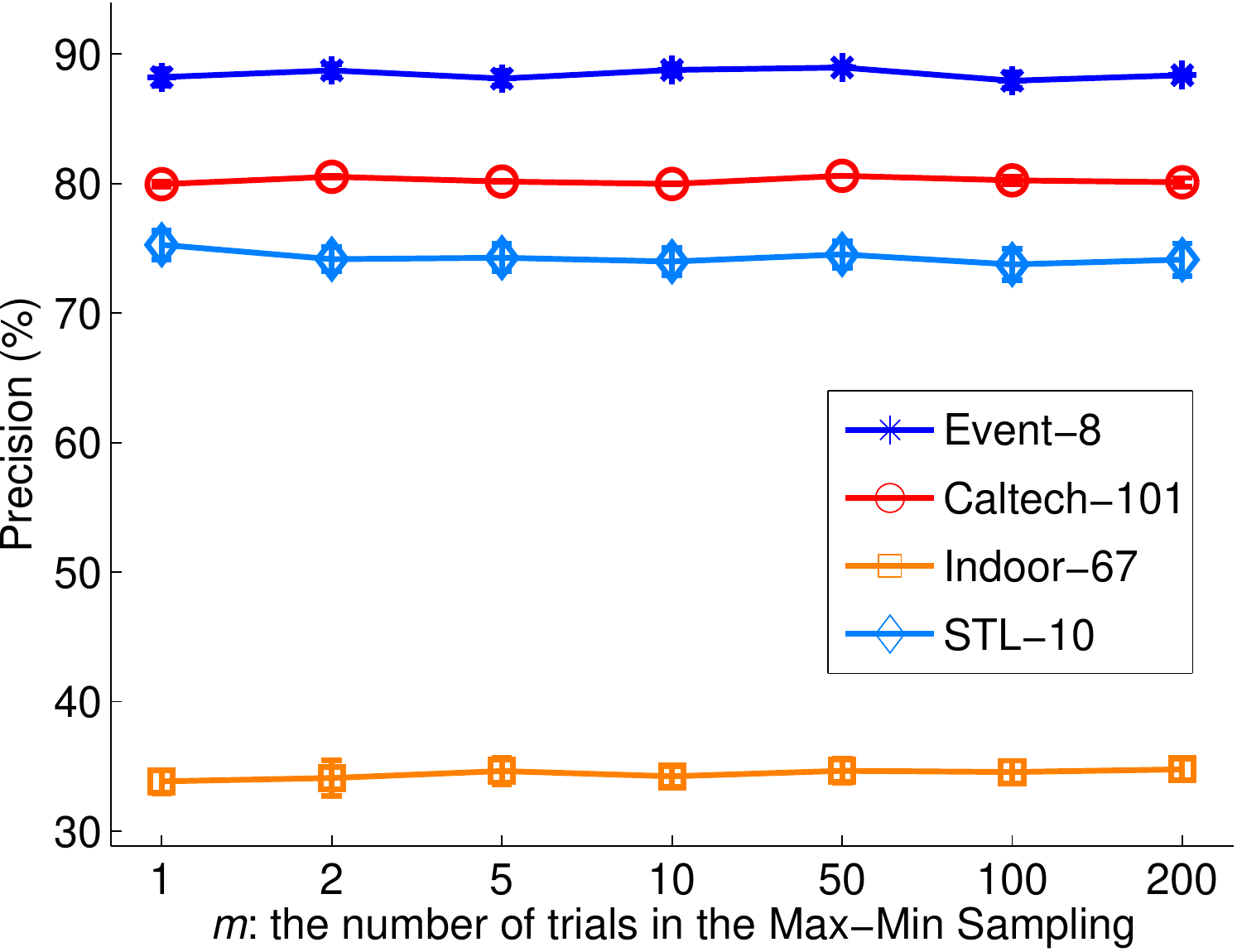} \\
\scriptsize{\text{(a) \emph{T}}} & \scriptsize{\text{ (b) \emph{r}}} & \scriptsize{\text{(c) \emph{n}}} & \scriptsize{\text{(d) \emph{m}}} \\
\end{array}$
\caption{Performance of EP as a function its parameters $T$,  $r$, $n$,
  and $m$, where LR is employed with $5$ labeled training images per class.}
  \label{fig:mrnt}
\end{figure*}

\subsubsection{Robustness to Parameters}
\label{sec:parameters}
In this section, we examine the influence of the parameters of our
method on its classification performance.  They are the total number of
prototype sets $T$, the number of prototypes in each set $r$, the
number of images in each prototype $n$, and the number of skeleton hypotheses
$m$ used in Max-Min Sampling. LR was used as the
classifier here. The parameters were evaluated as follows.
Each time the value of one changes while the other ones being kept fixed to the
values described in the experimental settings.

Figure~\ref{fig:mrnt} shows the results over a range of their
values. The figure shows that the performance of our method increases
pretty fast with $T$, but then stabilizes quickly. It implies that the
method benefits from exploiting more ``novel'' visual attributes
(image prototypes). After $T$ increases to some threshold (\eg $50$
for the eight datasets), basically no new attributes are added, and performance stops going up much. 
For $r$, the
figure shows that the performance generally increases with it. This is
expected because a large $r$ leads to precise attribute
assignment. In other words, a large $r$ generates more
prototypes per set, thus increasing the possibility of linking every
image to its desirable attribute.  However, we seen that when $r$
outpaces $24$, the increase is not worth the computing time. A larger
$r$ would lead to confusing attributes, as it starts to draw very
similar or even identical samples into different prototypes.  Also, a
large $r$ results in high-dimensional features, which in turn
cause over-fitting. 

For $n$, a similar trend was obtained -- as $n$ increases, the
characteristics of the prototypes are enriched, thus boosting the
performance. But beyond some threshold (\eg $10$ in our experiments),
more noisy images are introduced, thus degrading the performance. One
possible solution to further enrich the training samples of each
prototype is to perform image transformations such as
\emph{translation}, \emph{rotation}, and \emph{scaling} to the seed
images, and to add the transformed images into the prototype. This
technique of enriching training data has been successfully used
recently for image classification~\citep{transformation:cvpr14} and for feature learning~\citep{cnnfet14}.  For $m$, Figure~\ref{fig:mrnt} shows
that it does not affect the performance as much as the three parameters
analyzed so far. This does not mean that there is no need to use the
\emph{exotic-consistency} assumption. Instead, it suggests that a
random selection of $r$ images from a dataset of $l+u$ images already
fulfills the requirement of the assumption: images should be apart
from each other.  This is generally true because $r << l+u$ holds for
the datasets considered.


Although the performance of EP will be affected by the choice of its
parameters, we can see from Figure~\ref{fig:mrnt} that each of the
parameters has a wide range of reasonable values to choose from. It is
not difficult to choose a set of parameter values that produces better
results than competing methods (\cf Figure~\ref{fig:mrnt} and
Table~\ref{table:precision}).  Also, the parameters are quite
intuitive and their roles are similar to the parameters of some
other methods: analogues of $m$, $n$ and $T$ can be found in
RANSAC, $k$-NN, and Bagging, for instance.


\begin{figure*} 
  \centering
   $ \begin{array}{cccc}
\hspace{-2mm}
\includegraphics[width=0.35\linewidth, height=46mm]{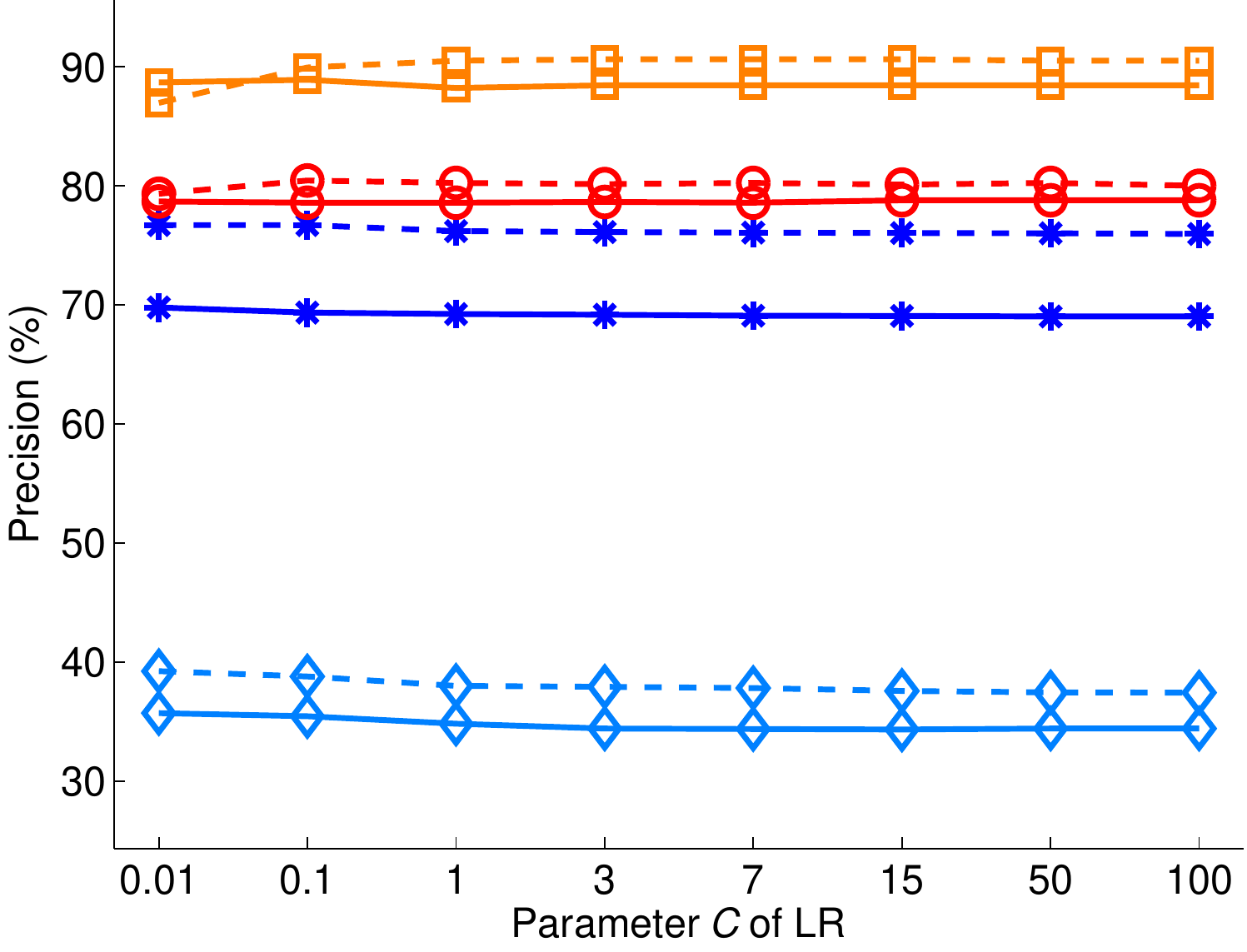} &  \hspace{-4mm}
\includegraphics[width=0.12\linewidth, height=42mm]{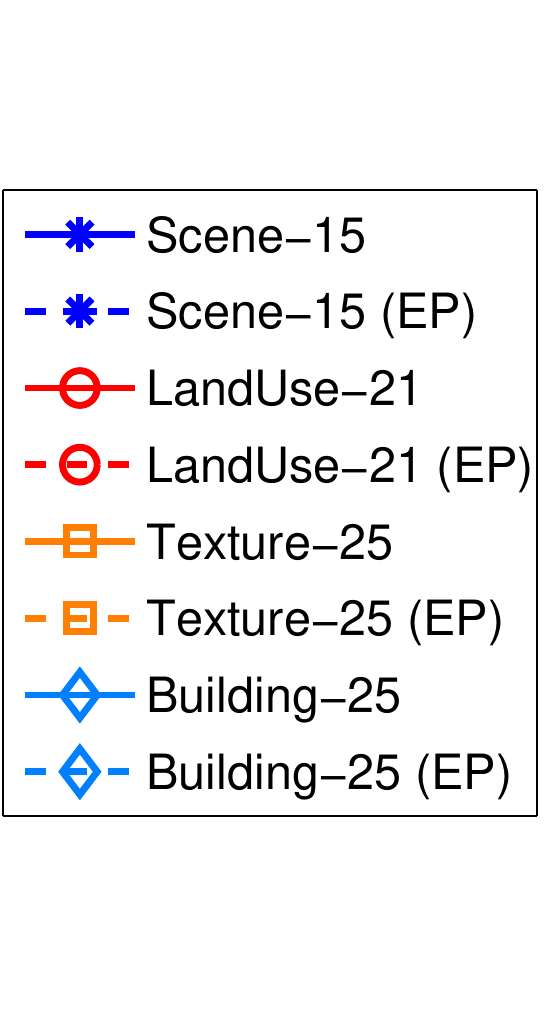} &  
\includegraphics[width=0.35\linewidth, height=46mm]{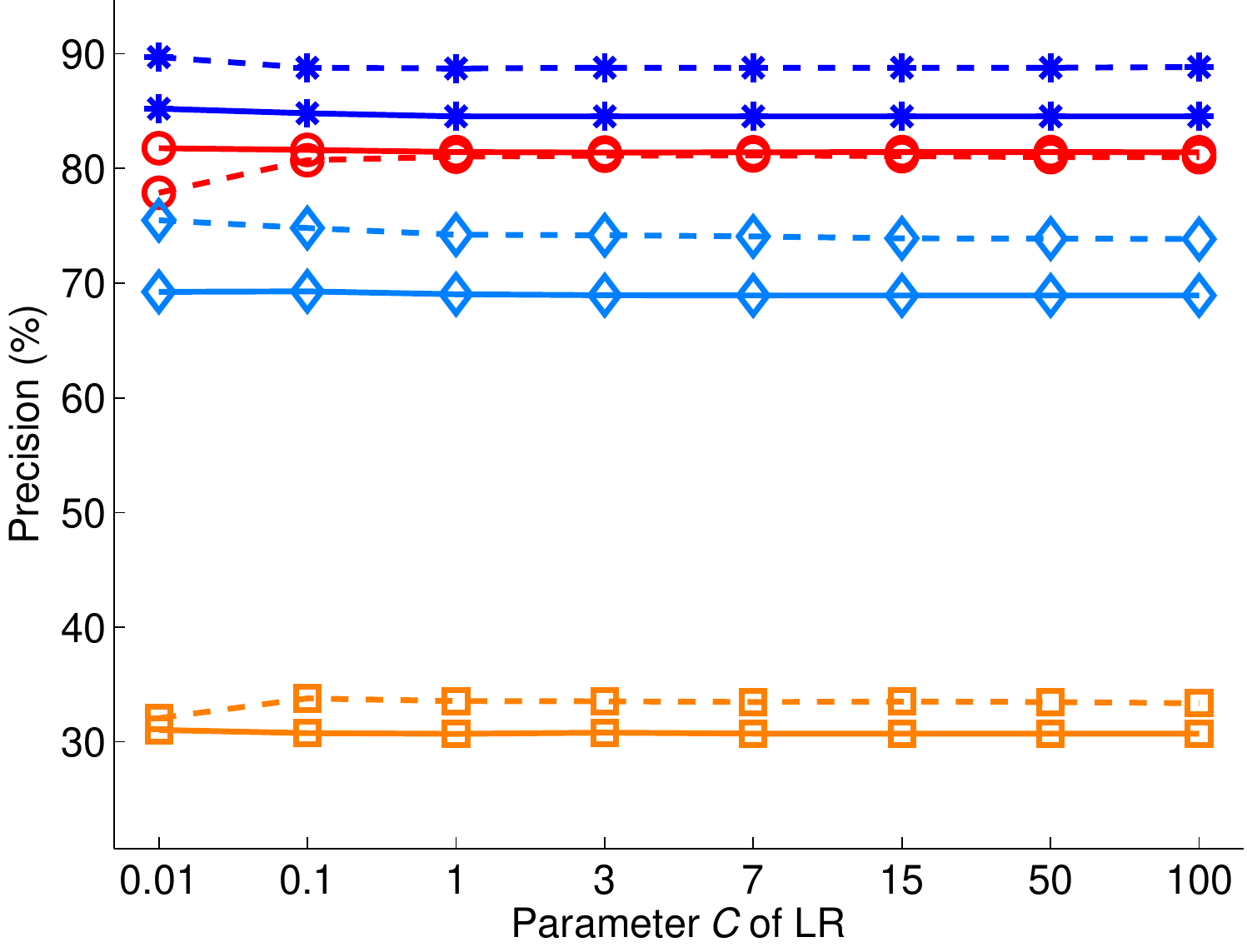} &  \hspace{-4mm}
\includegraphics[width=0.12\linewidth, height=42mm]{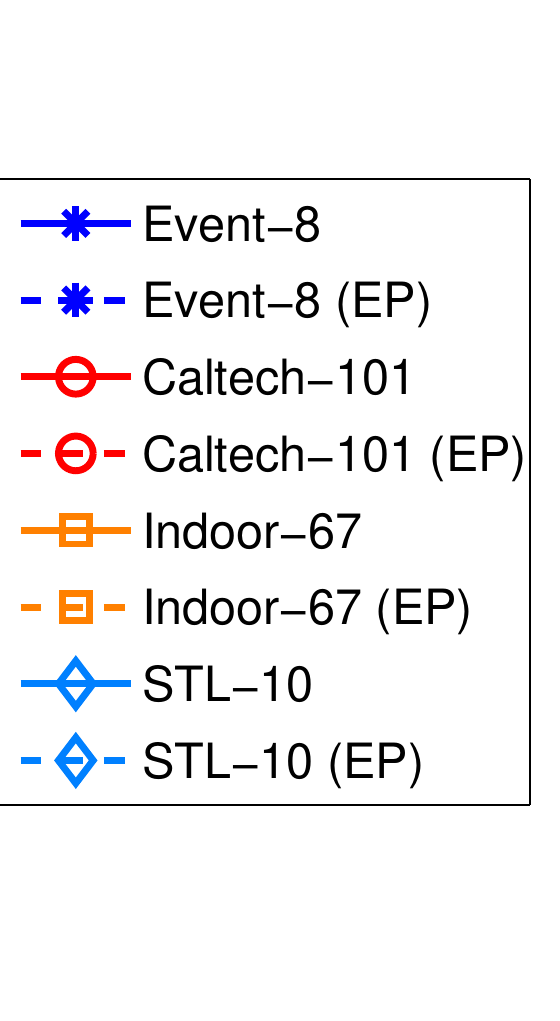} \\
\end{array}$
\caption{Comparison of our learned feature to the CNN feature~\cite{deep:bmvc14}, with different LR models.}
  \label{fig:classifiers}
\end{figure*}



\subsubsection{Robustness to Classifier Models}
In this section, we evaluate the robustness of our learned features
against classifier models. Different values of the balancing parameter
$C$ between model accuracy and model complexity were tested for the LR
classifier across the eight datasets.  $5$ labeled training examples
per class were used.  A set of values $\{0.01, 0.1, 1, 5, 15, 50,
100\}$ were tested for the parameter $C$ of LR. Figure~\ref{fig:classifiers}
shows the results. It is evident from the figure that our learned
feature consistently outperforms the original CNN feature over a large
range of parameter values for the classifier models. This property is
important for semi-supervised classification, as labeled data is
limited in this scenario and probably cannot afford model selection
techniques such as Cross-Validation.

 \subsubsection{Efficiency}
 Although additional time is needed for feature learning (the direct use of
 the original feature needs no training at this stage), our method is efficient. The efficiency is due to two reasons: 1) Training
 logistic regression is very efficient; and 2) the performance of our
 method stabilizes quickly with respect to $T$ as Figure\ref{fig:mrnt}
 shows. The training on the datasets takes $2-6$ minutes on a Core i5
 2.80 GHz desktop PC. Furthermore, our method is inherently
 parallelizable and can take advantage of multi-core processors. It is
 worth noting that this extra-training time is compensated by using a
 simpler classifier such as logistic regression for the
 classification.


\begin{figure*} 
  \centering
   $ \begin{array}{cccc}
\includegraphics[width=0.31\linewidth, height=40mm]{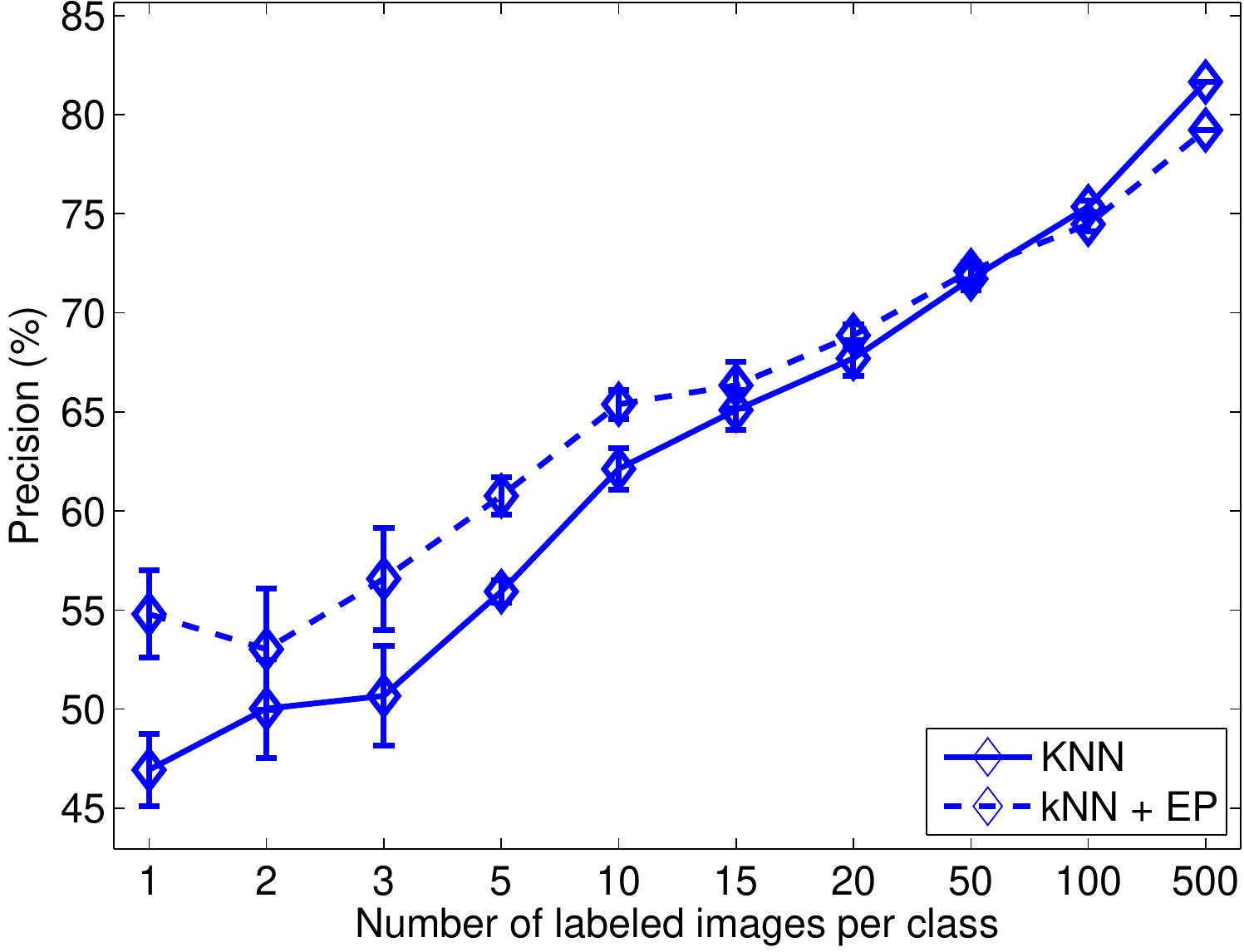}& 
\includegraphics[width=0.31\linewidth, height=40mm]{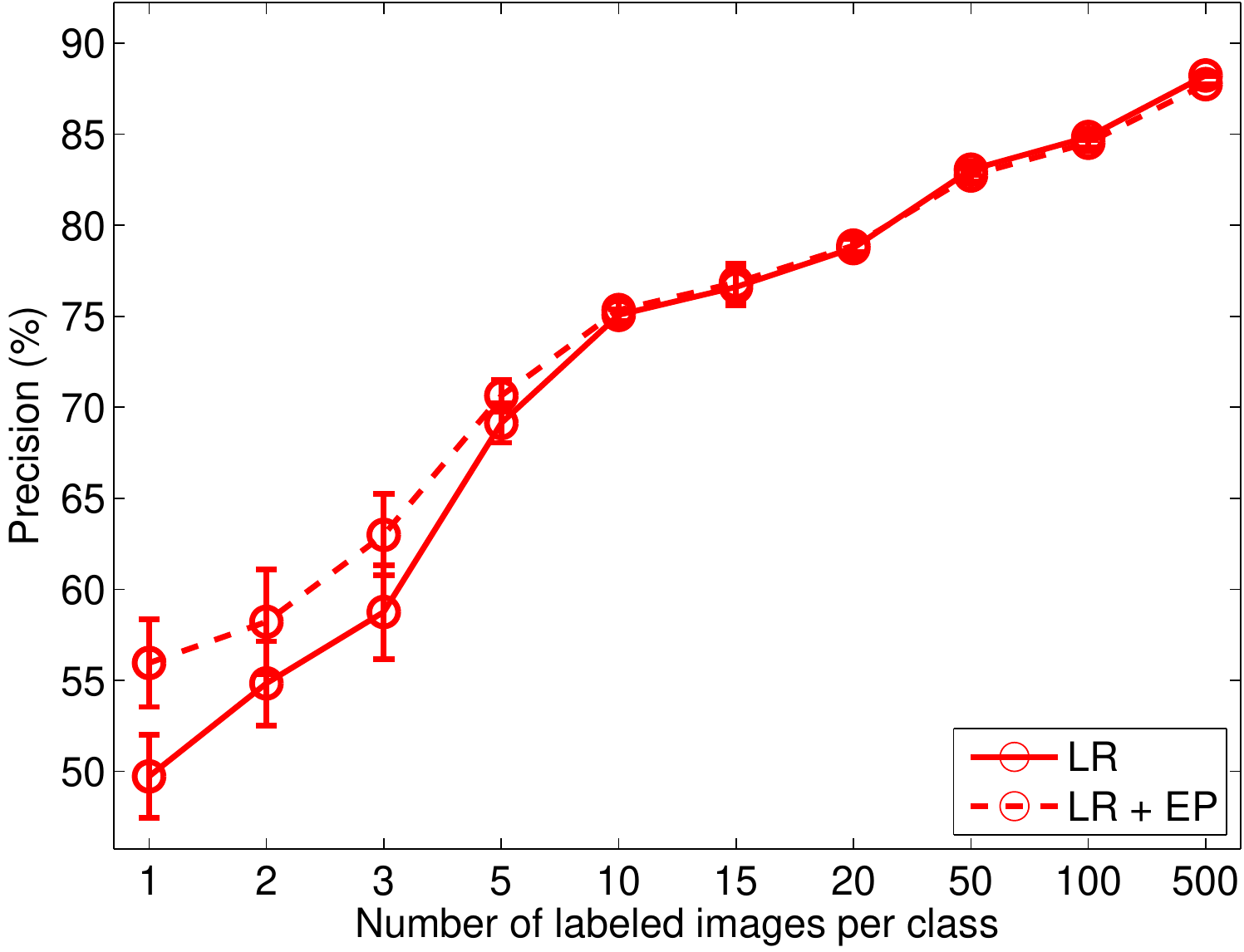}& 
\includegraphics[width=0.31\linewidth, height=40mm]{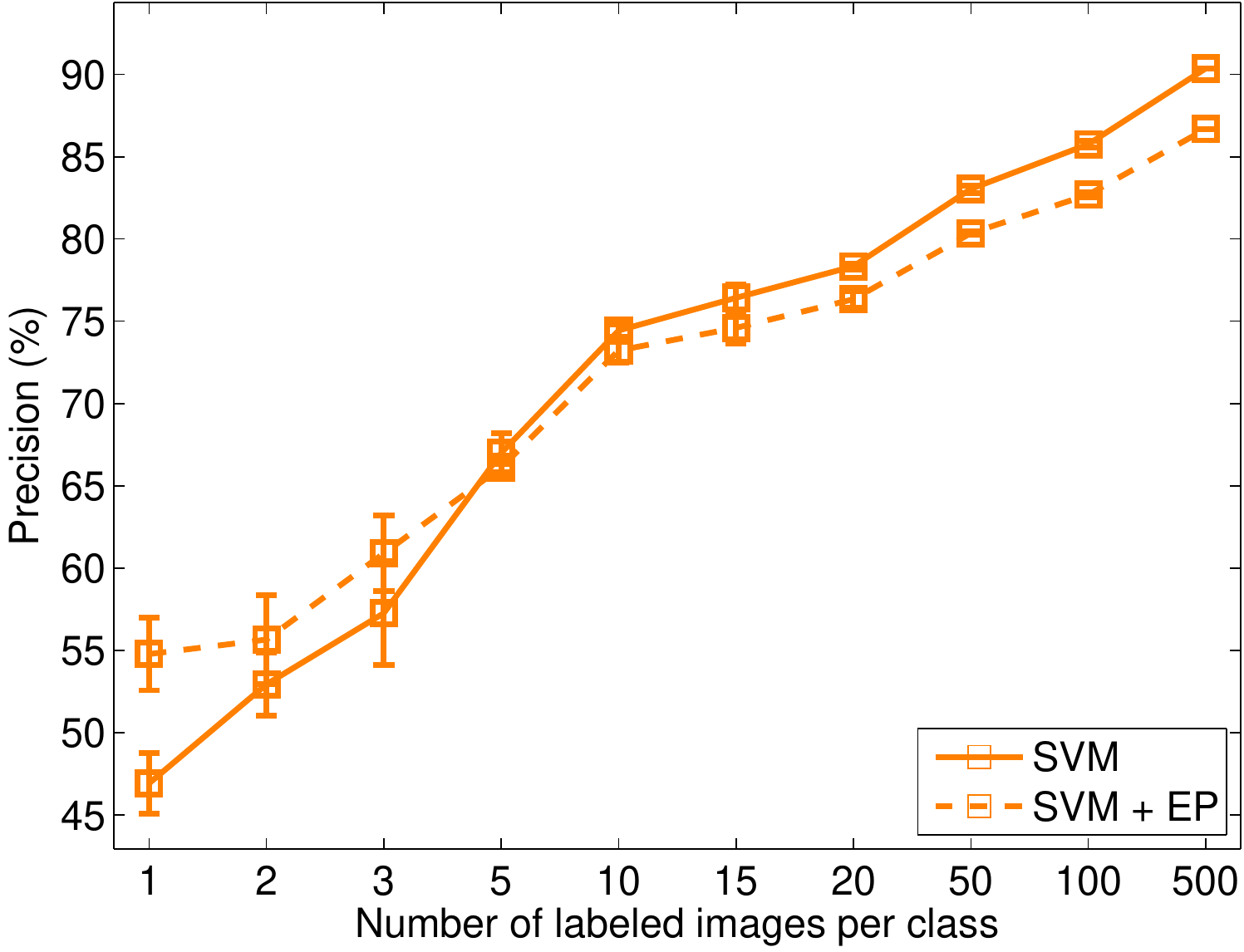} \\
\scriptsize{\text{(a) $k$-NN}}  & \scriptsize{\text{(b) LR}} & \scriptsize{\text{(c) SVMs}}
\end{array}$
\caption{Self-taught classification results on dataset STL-10, where EP is learned from the unlabeled images. The
  classifiers were tested with deep features, and our learned feature from it (indicated by ``+ EP").}
  \label{fig:stl}
\end{figure*}

\subsection{Self-taught Image Classification}
\label{sec:self}
In order to evaluate the generality of our method, we tested it in a
more general scenario, where the unlabeled data is the set of
$100,000$ unlabeled images from the STL-10 dataset. Projection
functions were learned from this unlabeled dataset and the performance
was tested on the STL-10 dataset. Again, we held the training image
and test images as a whole, and chose only a small fraction as
training images (for the classifiers) with others as test images for
evaluation.  The average accuracy of $5$ runs with random
training-test splits was reported.  Figure~\ref{fig:stl} shows the
classification performance with different numbers of labeled training
images per class.  From the figure and table, it can be observed that our
learned feature from the random image collection still outperforms the
original CNN feature when the number of labeled training images is
small.
This is a very helpful property for semi-supervised learning,
as it happens quite often that one has no prior access to the data to
be classified. The success could be ascribed to the fact that the
``universal visual world'' (the random image collection) contains
abundant high-level, valuable visual attributes such as ``blue and
open'' in some image clusters and ``textured and man-made'' in
others. Exploiting these ``hidden'' visual attributes is very
beneficial for narrowing down the semantic gap between low-level
features and high-level classification tasks.

However, the figure also shows that as the number of labeled training
images increases, the advantage of our learned feature vanishes. The
method even produces worse results than the original CNN feature when
the number of training samples is large.  This is to be expected as
the method is designed to improve classification systems by exploiting
unlabeled data. Therefore, when a sufficient
number of labeled images are available, introducing additional
unlabeled ones may hurt the system.  This is a general, open problem
for semi-supervised learning (self-taught
learning)~\citep{neverhurt:icml11}. One possible solution is to study
when the classification systems should switch from semi-supervised
learning to fully supervised learning. Another solution could be to
use the labeled training images directly as the skeleton to generate
the prototype sets. This strategy, however, is more limited than
ours, and is difficult to use for tasks, such as image
clustering, where no labeled samples are
available. We leave these issues as future work. 







\subsection{Image Clustering}

\begin{table*}[tb] \small
\setlength\tabcolsep{0.18em} {
$\begin{array}{ccccccccccccc}

\begin{tabular}{cc}
\toprule
\multicolumn{1}{c}{\footnotesize{Methods}} \\ \midrule
$k$-means \\  
$k$-means \bf{+ EP} \\    
\footnotesize{Spectral Clustering} \\
\footnotesize{Spectral Clustering \bf{+ EP}} \\
\bottomrule
\end{tabular}

\begin{tabular}{cc}
\toprule
\multicolumn{1}{c}{\footnotesize{Scene-15}} \\ \midrule
65.5  \\
\underline{71.5}  \\  
69.6  \\
\bf{73.6}  \\
\bottomrule
\end{tabular}

\begin{tabular}{cc}
\toprule
\multicolumn{1}{c}{\footnotesize{LandUse-21}} \\ \midrule
56.3  \\
\underline{63.6}  \\   
59.8  \\
\bf{65.2}  \\
\bottomrule
\end{tabular}

\begin{tabular}{cc}
\toprule
\multicolumn{1}{c}{\footnotesize{Texture-25}} \\ \midrule
59.2  \\
\bf{73.1}  \\ 
66.6  \\
\underline{70.1}  \\
\bottomrule
\end{tabular}

\begin{tabular}{cc}
\toprule
\multicolumn{1}{c}{\footnotesize{Building-25}} \\ \midrule
39.3  \\
\bf{43.8}  \\ 
33.3  \\
\underline{41.0}  \\
\bottomrule 
\end{tabular}

\begin{tabular}{cc}
\toprule
\multicolumn{1}{c}{\footnotesize{Event-8}} \\ \midrule
76.7  \\
\bf{87.3}  \\ 
82.7  \\
\underline{86.5}  \\  
\bottomrule
\end{tabular}

\begin{tabular}{cc}
\toprule
\multicolumn{1}{c}{\footnotesize{Caltech-101}} \\ \midrule
67.7  \\
\underline{69.4}  \\  
68.2  \\
\bf{70.7}  \\
\bottomrule 
\end{tabular}

\begin{tabular}{cc}
\toprule
\multicolumn{1}{c}{\footnotesize{Indoor-67}} \\ \midrule
33.1  \\
\underline{37.0}  \\  
31.5  \\
\bf{37.2}  \\
\bottomrule
\end{tabular}

\begin{tabular}{cc}
\toprule
\multicolumn{1}{c}{\footnotesize{STL-10}} \\ \midrule
57.0  \\
\underline{63.5}  \\  
52.8  \\
\bf{66.4}  \\
\bottomrule
\end{tabular}


\end{array} $
\vspace{1mm}
\centering 
\caption{Purity (\%) of image clustering on the eight datasets, where the CNN feature \citep{deep:bmvc14} and our learned feature from it (indicated by + EP) are used.  The best results are indicated in \textbf{bold},
  and the second best is \underline{underlined}.} 
\label{table:clustering}}
\end{table*}

In this section, we evaluated our learned feature for the task of
image clustering. Given a collection of images without any labels, the
task is to group them so that images in the same group are more
(semantically) similar to each other than to those in other groups.
We follow existing work \citep{Sivic05b,
  Tuytelaars_UnsupervisedSurvey, dai, dai:eccv12b, fakton:eccv12} and
evaluate the task on the image classification datasets, in particular 
on the eight datasets used for semi-supervised image
classification.  To the best of our knowledge, we are the first
to evaluate the performance of image clustering on as many as eight
standard classification datasets, some of which are still very
challenging for supervised image classification. 
Most clustering methods have been tested only on relatively simple
datasets, such as $4$, $7$ and $20$ classes of the Caltech dataset,
and $5$ classes of the ETH shape dataset.

Since our main aim is to validate whether the proposed learning is
able to boost the performance of the original feature for image
clustering, we chose two standard clustering algorithms -- Spectral
Clustering and $k$-means -- to compare the two features.  As to the
implementation, we use the parallel implementation of
\citep{parallel:sc} for Spectral Clustering and the vl-feat library of
\citep{vlfeat} for $k$-means algorithm.  Since Spectral
Clustering and $k$-means both require the number of clusters
as a parameter, we set it to the number of semantic classes of the datasets, leading to
weakly-supervised image clustering. 

Table \ref{table:clustering} lists the results of the two features
when combined with $k$-means and Spectral Clustering.  Purity is used
as the evaluation criterion, which measures the percentage of images
from the dominant class within their clusters, averaged over all
clusters. The dominant class of a cluster is the (semantic) class that
has more image members than other classes in the cluster.  It is easy
to see from the table that features learned by EP outperform the
original CNN features for image clustering by a considerable
margin. For instance, when $k$-means is used, EP outperforms the CNN
feature by $9.6\%$ on Event-8, and by $6.5\%$ on STL-10; when Spectral
Clustering is used, the improvement is $4.0\%$ on Scene-15, and
$5.7\%$ on Indoor-67.  Again, our feature is learned from the original
CNN feature, but goes beyond one single image and captures the
\emph{similarity} relationship among images. The superior performance
of the learned feature suggests that it is worth some effort to
analyze properties of the datasets to learn a better feature
representation before performing image clustering. This is useful for the task of clustering, as all the data is
available to use from the very beginning. This pre-processing step of
analyzing datasets has not yet raised much attention in the community. We
hope that this work will stimulate more efforts in this direction.

\section{Conclusion} 
\label{sec:conclusion}
This paper has tackled the problem of feature learning for the tasks of
semi-supervised image classification and image clustering.  We
proposed a simple, yet effective feature learning method to exploit the available,
unlabeled data. By using two consistency assumptions, we generate a diverse set of
training data for surrogate classes to learn visual attributes in a
discriminative way.  By doing so, images are classified and linked to
the surrogate classes -- images are represented with their affinities
to a rich set of discovered image attributes for classification and
clustering. Experiments on eight datasets showed the superior
performance of the learned feature for both semi-supervised image
classification and image clustering. In addition, the method is
conceptually simple, computationally efficient, and flexible to use.
The future work is to extend the method to relevant tasks, such as image segmentation.



\noindent
\textbf{Acknowledgements.} The work is supported by the ERC Advanced Grant Varcity (\#273940).


{
\small
\bibliographystyle{elsarticle-num}
\bibliography{egbib}
}
\end{document}